\documentclass[conference]{IEEEtran}
\IEEEoverridecommandlockouts
\usepackage{cite}
\usepackage{amsmath,amssymb,amsfonts}
\usepackage{algorithm}
\usepackage{algorithmic}
\usepackage{graphicx}
\usepackage{textcomp}
\usepackage{soul}
\usepackage{xcolor}
\usepackage{multirow}
\usepackage{subfigure}
\usepackage{url}

\def\BibTeX{{\rm B\kern-.05em{\sc i\kern-.025em b}\kern-.08em
    T\kern-.1667em\lower.7ex\hbox{E}\kern-.125emX}}

\usepackage{array}
\newcolumntype{L}[1]{>{\raggedright\let\newline\\\arraybackslash\hspace{0pt}}m{#1}}
\newcolumntype{C}[1]{>{\centering\let\newline  \\\arraybackslash\hspace{0pt}}m{#1}}
\newcolumntype{R}[1]{>{\raggedleft\let\newline \\\arraybackslash\hspace{0pt}}m{#1}}

\begin{document}

\title{NSCaching: Simple and Efficient Negative Sampling for Knowledge Graph Embedding
\thanks{This work is partially done when Y. Zhang was an intern in 4Paradigm Inc,
	and Q. Yao is the correspondence author.}
}

\author{\IEEEauthorblockN{Yongqi Zhang\IEEEauthorrefmark{2},
		Quanming Yao\IEEEauthorrefmark{3}, Yingxia Shao\IEEEauthorrefmark{4} and
		Lei Chen\IEEEauthorrefmark{2}}
	\IEEEauthorblockA{\IEEEauthorrefmark{2}The Hong Kong University of Science and Technology, Hong Kong SAR, China\\
		\IEEEauthorrefmark{3}4Paradigm Inc., Beijing, China\\
		\IEEEauthorrefmark{4}Beijing University of Posts and Telecommunications, Beijing, China\\
		\IEEEauthorrefmark{2}\{yzhangee,leichen\}@cse.ust.hk,
		\IEEEauthorrefmark{3}yaoquanming@4paradigm.com,
		\IEEEauthorrefmark{4}shaoyx@bupt.edu.cn}}

\maketitle



\begin{abstract}
Knowledge graph (KG) embedding is a fundamental problem in data mining research with many real-world applications.
It aims to encode the entities and relations in the graph into low dimensional vector space,
which can be used for subsequent algorithms.
Negative sampling,
which samples negative triplets from non-observed ones in the training data, 
is an important step in KG embedding.
Recently,
generative adversarial network (GAN),
has been introduced in negative sampling.
By sampling negative triplets with large scores,
these methods avoid the problem of vanishing gradient and thus obtain better performance.
However,
using GAN makes the original model more complex and hard to train,
where reinforcement learning must be used.
In this paper,
motivated by the observation that negative triplets with large scores are important but rare,
we propose to directly keep track of them with cache.
However,
how to sample from and update the cache are two important questions.
We carefully design the solutions,
which are not only efficient but also achieve good balance between exploration and exploitation.
In this way,
our method acts as a ``distilled'' version of previous GAN-based methods,
which does not waste training time on additional parameters to fit the full distribution of negative triplets.
The extensive experiments show that 
our method can gain significant improvement on various KG embedding models, 
and outperform the state-of-the-arts negative sampling methods based on GAN.
\end{abstract}


\section{Introduction}

Knowledge graph (KG) is a special kind of graph structure, 
with entities as nodes, 
and relations as directed edges. 
Each edge (also called a fact) is represented as a triplet with the form \emph{(head entity, relation, tail entity)}, 
which is denoted as $(h, r, t)$, indicating that two entities are connected by a specific relation, 
e.g. \emph{(Shakespeare, isAuthorOf,  Hamlet)} \cite{bollacker2008freebase,miller1995wordnet,auer2007dbpedia,suchanek2007yago}.
KG is very general and useful,
and it has been used as fundamental building blocks for many applications,
e.g., structured search \cite{singhal2012introducing},
question answering \cite{bordes2014open, bordes2014question},
and intelligent virtual assistant \cite{white1991knowledge}.
Such importance
has also inspired many famous KG projects,
e.g.,
FreeBase \cite{bollacker2008freebase},
DBpedia \cite{auer2007dbpedia},
and YAGO \cite{suchanek2007yago}.

However,
as these triplets are hard to manipulate,
one fundamental problem is how to find a good representation for entities and relations in the KG \cite{nickel2016review}.
Early works towards this goal lie in statistical relational learning using the symbolic triplet data \cite{kok2007statistical, getoor2007introduction, lao2011random}. 
However, these methods neither lead to good generalization performance, 
nor can they be applied for large scale knowledge graphs. 
Recently, 
graph embedding techniques \cite{wang2017knowledge} have been introduced in KG.
These methods attempt to encode entities and relations in KG into a low-dimensional vector space
while capturing nodes' and edges' connection properties.
They are scalable and have also shown a promising performance in basic KG tasks, 
such as link prediction and triplet classification \cite{bordes2013translating, wang2017knowledge}. 

{
Besides,
based on the learned entity and relation embeddings, downstream tasks, 
such as entity classification \cite{nickel2011three} and entity linking \cite{bordes2014semantic},
can also be benefited.
Given that the relation encoding entity types (denoted as \textit{IsA}) 
or the relation encoding equivalent entities (denoted as \textit{EqualTo}) is contained in the KG,
and has been included into the learning process,
entity classification can be treated as the link prediction task $(x, IsA, ?)$, 
and entity linking treated as triplet classification task $(a, EqualTo, b)$.
A more direct entity linking method proposed in \cite{nickel2011three} is to check the similarity score 
between embeddings of two entities. 
}

In recent years,
constructing new scoring functions which can better model the complex interactions between entities and relations 
have been the main focus for improving KG embedding's performance
\cite{wang2014knowledge,ji2015knowledge,yang2014embedding,trouillon2016complex}.
However,
another very important perspective of KG embedding,
i.e., negative sampling, is not sufficiently emphasized.
The need of negative sampling comes from the fact that there are only positive triplets in KG \cite{drumond2012predicting}.
To avoid trivial solutions of the embedding,
for each positive triplet, a set
that contains its all possible negative samples,
needs to be hand-made.
Then,
for the effectiveness and efficiency of stochastic updates in the KG embedding,
once we have picked up a positive triplet,
we also need to sample a negative triplet from its corresponding negative sample set.
Unfortunately,
the quality of these negative triplets does matter.

Due to its simplicity and efficiency,
uniform sampling is broadly used in KG embedding \cite{wang2017knowledge}.
However,
it is a fixed scheme and ignores changes on the distribution of negative triplets during the training.
Thus,
it suffers seriously from the \textit{vanishing gradient} problem.
Specifically,
as observed in \cite{wang2018incorporating},
most negative triplets in the sampling set are easily classified ones.
{Since scoring functions tend to give observed (positive) triplets large values,
	as training goes,
	scores (evaluated from scoring functions) for most non-observed (probably negative) triplets become smaller.
	Thus, 
	when negative triplets are uniformly sampled,
	it is very likely that we pick up one with zero gradient.
	As a result, 
	the training process of KG embedding 
	will be impeded by such vanishing gradients rather than by the optimization algorithm.
	Such problem prevents KG embedding getting desired performance.}
A better sampling scheme,
i.e.,
Bernoulli sampling, 
is introduced in \cite{wang2014knowledge}.
It improves uniform sampling
by considering one-to-many,
many-to-many,
and many-to-one mapping in relation between head and tail.
However,
it is still a fixed sampling scheme,
which suffers from vanishing gradients.

Therefore,
high-quality negative triplets should have large scores.
To efficiently capture them during training,
we have two main challenges for the negative sampling:
(i). How to capture and model negative triplets' dynamic distribution? 
and (ii). How can we sample negative triplets in an efficient way?
%
%
%
%
Recently,
there are two pioneered works, 
i.e.,
IGAN \cite{wang2018incorporating} and KBGAN \cite{cai2018kbgan},
attempting to address these challenges.
Their ideas are both replacing the fixed sampling scheme with
a generative adversarial network (GAN) \cite{goodfellow2014generative}.
{However,
		the GAN-based solutions still have many problems.
		First, GAN increases the number of training parameters because an extra generator is introduced.
		Second,
		GAN training can suffer from instability and degeneracy \cite{arjovsky2017wasserstein, gulrajani2017improved}, 
		and the REINFORCE gradient \cite{williams1992simple} used in IGAN and KBGAN is known to have high variance.
		These drawbacks lead to instable performance for different scoring functions, 
		and hence pretrain becomes a must for both IGAN and KBGAN.}

In this paper, to address the challenges of high-quality negative sampling 
while avoiding the problems from using GAN, we propose a new negative sampling method based on cache, 
called NSCaching.
With empirically studying the score distribution of negative samples, we find that the score distribution is highly skew,
i.e., there are only a few negative triplets with large scores and the rest are useless.
This observation motivates to only maintain high-quality negative triplets during the training, 
and dynamically update the maintained triplets. 
First, we store the high-quality negative triplets in cache, 
and then design importance sampling (IS) strategy to update the cache.
The IS strategy can not only capture the dynamic characteristic of the distribution, 
but also benefit the efficiency of NSCaching. 
Furthermore, we also take good care of ``exploration and exploitation'',
which balances exploring all possible high-quality negative triplets
and sampling from a few large score negative triplets in cache. 
Contributions of our work are summarized as follows:
\begin{itemize}
	\item We propose a simple and efficient negative sampling scheme, 
	NSCaching. It is a general negative sampling scheme, 
	which can be injected into all popularly used KG embedding models. 
	NSCaching has fewer parameters than both IGAN and KBGAN, 
	and can be trained with gradient descent as the original KG embedding models.
	
	\item We propose the uniform strategy to sample from the cache
	and IS strategy to update the cache in NSCaching 
	with good care of \textit{``exploration and exploitation''}.
	
	
	\item We analyze the connection between NSCaching and the \textit{self-paced learning} \cite{kumar2010self,bengio2009curriculum}.
	We show NSCaching can firstly learn easily classified samples, 
	and then gradually switch to harder samples.
	
	\item 
	We conduct experiments on four popular data sets, i.e.,  WN18 and FB15K, and their variants WN18RR and FB15K237. 
	Experimental results demonstrate that our method is very efficient and is more effective than the state-of-the-arts, 
	i.e., IGAN and KBGAN, as well.
\end{itemize}

\noindent
\textbf{Notation.}
We denote the set of entities as $\mathcal E$ and set of relations as $\mathcal R$.  
A fact (edge) in KG is represented by a triplet,
i.e., $(h, r, t)$,
where $h \in \mathcal{E}$ is the head entity,
$t \in \mathcal{E}$ is the tail entity,
and $r \in \mathcal{R}$ is the relationship.
Observed facts in a KG are represented by a set $\mathcal S \equiv \{ (h, r, t) \}$.
Finally,
we denote the embedding vectors of $h$, $r$ and $t$ by its corresponding boldface character, 
i.e. $\mathbf h$,~$\mathbf r$ and $\mathbf t$.  

%
%
%

\section{Preliminary:
Framework of KG Embedding}
\label{sec-prelim}

%
%
%


Here,
we first introduce the general framework 
for training KG embedding models in Section~\ref{sec:uniframe}.
Then,
we describe its two key components,
i.e., 
negative sampling and scoring function
in Section~\ref{sec:negasamp} and \ref{sec:scorefunc} respectively.

\subsection{The General Framework}
\label{sec:uniframe}


To build a KG embedding model,
the most important thing is to design a scoring function $f$,
which captures the interactions between two entities based on a relation \cite{wang2017knowledge}.
Different scoring functions have their own weaknesses and strengths in capturing the underneath interactions.
{
Besides,
the observed facts in KG are supposed to have larger scores than non-observed ones. 
With the factual information, 
the embeddings are learned by solving the optimization problem that maximizes 
the scoring function for positive triplets and minimizes for non-observed triplets
at the same time.
}
Based on the properties of scoring functions,
KG embedding models are generally divided into two categories.
The first one is translational distance model, i.e.,
\begin{equation}
L(\mathcal{E}, \mathcal{R}) 
= \!\!\!\!\!
\sum_{(h,r,t)\in\mathcal S} 
\!\!\!\!\!
\left[\gamma -f(h,r,t) + f(\bar{h},r,\bar{t})\right]_+,
\label{eq:distance}
\end{equation}
and the second one is semantic matching model, i.e.,
\begin{equation}
L(\mathcal{E}, \mathcal{R})  
= \!\!\!\!\!
\sum_{(h, r, t)\in\mathcal S} \!\!\!\!\!
\left[ 
\ell\left( +1, f(h,r,t) \right) 
+ \ell\left( -1, f(\bar{h},r,\bar{t}) \right)
\right],
\label{eq:sematic}
\end{equation}
where
$(\bar{h},r,\bar{t}) \not\in \mathcal{S}$ is the hand-made negative triplet for $(h,r,t)$ and
$\ell(\alpha, \beta) = \log\left( 1 + \exp( - \alpha \beta ) \right)$ is the logistic loss.

\begin{algorithm}[ht]
	\caption{General framework of KG embedding.}
	\label{alg:basic}
	\begin{algorithmic}[1]
		\REQUIRE training set $\mathcal{S} = \{(h, r, t)\}$, embedding dimension $d$
		and scoring function $f$;
		
		\STATE initialize the embeddings for each $e \in \mathcal E$ and $r \in \mathcal R$.
		\FOR{$i = 1, \cdots, T$}
		
		\STATE sample a mini-batch $\mathcal S_{\text{batch}} \in \mathcal{S}$ of size $m$;
		\FOR{each $(h, r, t)\in\mathcal S_{\text{batch}}$}
		
		\STATE sample a negative triplet $(\bar{h}, r, \bar{t}) \in \bar{\mathcal{S}}_{(h,r,t)}$;
		\\ // \textit{negative sampling}

		\STATE update parameters of embeddings w.r.t. the gradients
		using (i). translational distance models:
		\begin{align}
		\nabla \left[\gamma -f\left( h,r,t \right) + f\left( \bar{h},r,\bar{t}\right) \right]_+,
		\label{eq:graddist}
		\end{align}
		or (ii). semantic matching models:
		\begin{align}
		\nabla \ell\left( +1, f(h,r,t) \right) + \nabla \ell\left( -1, f(\bar{h}, r, \bar{t}) \right);
		\label{eq:gradsema}
		\end{align}
		
		\ENDFOR
		
		\ENDFOR
	\end{algorithmic}
\end{algorithm}

The above two objectives can be optimized by
using stochastic gradient descent
in an unified framework (Algorithm~\ref{alg:basic}).
In each iteration,
a mini-batch $\mathcal{S}_{\text{batch}}$ of size $m$ is firstly sampled from $\mathcal{S}$ at step~3.  
In step~5,
since there are no negative triplets in $\mathcal{S}$,
a set $\bar{\mathcal{S}}_{(h,r,t)}$, i.e.,
\begin{align}
\!\!\!\!
\bar{\mathcal{S}}_{(h, r, t)} = 
\left\lbrace  
(\bar{h}, r, t)\notin \mathcal{S}
\,|\, 
\bar{h}\in\mathcal E \right\rbrace  
\cup 
\left\lbrace (h, r, \bar{t})\notin \mathcal{S}
\,|\,
\bar{t} \in \mathcal E
\right\rbrace
\!,\!\!
\label{eq:nega}
\end{align}
which contains negative triplets for $(h,r,t)$,
is made,
and one negative triplet $(\bar{h}, r, \bar{t})$ is sampled from $\bar{\mathcal{S}}_{(h,r,t)}$.
Finally,
embedding parameters are updated in step~6.
Thus,
in optimization,
the most important problem is how to do negative sampling,
i.e. generate and sample negative triplet from $\bar{\mathcal{S}}_{(h,r,t)}$.

\subsection{Negative Sampling}
\label{sec:negasamp}


Existing works on negative sampling can be divided into two categories,
i.e., sample from fixed and sample from dynamic distributions.

\subsubsection{Sample from fixed distributions}

In the early works \cite{bordes2013translating}, 
negative triplets are \emph{uniformly} sampled from the set $\bar{\mathcal{S}}_{(h, r, t)}$.
Such strategy is simple and efficient.
Later,
a better sampling scheme,
i.e.,
Bernoulli sampling, 
is introduced in \cite{wang2014knowledge}.
It improves uniform sampling
by reducing the appearance of false negative triplets
existing in one-to-many,
many-to-many,
and many-to-one relations between head and tail entities.
However,
as mentioned in the introduction,
they still sample from fixed distributions,
which can neither model the dynamic changes in distributions of negative triplets
nor sample triplets with large scores.
Thus,
they seriously suffer from vanishing gradient.

\subsubsection{Sample from dynamic distributions}


More recently, 
two pioneered works \cite{wang2018incorporating,cai2018kbgan}
made a more dedicated analysis of problems with fixed sampling scheme.
They observed that most of the negative triplets are easy ones,
of which scores quickly go small during the training.
This leads to the vanishing gradient problem
if a fixed sampling scheme is used.
Motivated by the success of Generative Adversarial Network (GAN) \cite{goodfellow2014generative} and its ability to model dynamic distribution,
IGAN and KBGAN introduce GAN for negative sampling in KG.

When GAN is applied to negative sampling, a jointly trained generator serves as a sampler 
that can not only generate high-quality triplets by confusing the discriminator, 
but also dynamically adapt to the new distributions by keeping training.
The discriminator, i.e., the KG embedding model,
learns to distinguish between the positive triplets and the negative triplets
selected by the generator.
Under an alternating training procedure,
the generator dynamically approximates the negative sample distribution,
and the KG embedding model is improved by high-quality negative samples.

Specifically,
given a positive triplet $(h, r, t)$, 
IGAN models the distribution $\bar{h}, \bar{t}\sim p(e|(h, r, t))$ over all entities 
to form a negative triplet $(\bar{h}, r, \bar{t})$.
The quality of $(\bar{h}, r, \bar{t})$ is measured by the scoring function of the discriminator,
i.e. the target KG embedding model.
By joint training, IGAN can dynamically sample negative triplets with high quality.
KBGAN operates in a different way.
Instead of modeling a distribution over the whole entity set,
KBGAN learns to sample from a subset of random entities.
Namely, it first uniformly samples a set of entities to form a candidate set $\mathcal Neg = \{(\bar{h}, r, \bar{t}) \}$,
and then picks up one triplet from it.
Under the framework of GAN, 
generator in KBGAN can approximate the score distribution of triplets in the set $\mathcal Neg$,
and sample a triplet with relatively high quality.

{
Even though GAN provides a solution to model the dynamic negative sample distribution,
it is famous for suffering from instability and degeneracy \cite{arjovsky2017wasserstein,gulrajani2017improved}.
Besides, REINFORCE gradient \cite{williams1992simple} has to be used, which is known to have high variance.
Thus, 
pretrain is a must for both IGAN and KBGAN.
Finally, it increases the number of model's parameters and brings extra costs on training.
}

\subsection{Scoring Functions}
\label{sec:scorefunc}

The design of scoring function has been the main power source
for improving embedding performance in recent years.
Depending on the property of scoring functions,
they are used in either translational distance 
or semantic matching models.

\subsubsection{Translational distance model}

The simplest and most representative \emph{translational distance model} is TransE \cite{bordes2013translating}. 
Inspired from the word representation learning area \cite{mikolov2013linguistic}, 
if a triplet $(h, r, t)$ is true, the entity embeddings $\mathbf h, \mathbf t$ should be connected by the relational vector $\mathbf r$, 
i.e. $\mathbf h+\mathbf r\approx \mathbf t$. 
Under this assumption for example, two facts (\emph{China, Capital, Beijing}) and (\emph{UK, Capital, London}) will enjoy a relation that $China - UK \approx Beijing - London$ in the embedding space. 
Thus in TransE, 
the scoring function is defined as the negative translational distance of $\mathbf h$ and $\mathbf t$ connected by relation $\mathbf r$,
i.e., $f(h, r, t) = \left\|\mathbf h + \mathbf r -\mathbf t\right\|_1$.

Despite the simplicity of TransE, it faces the problem when dealing with one-to-many, many-to-one and many-to-many relations. 
Take one-to-many relation for example, TransE enforces $\mathbf h+\mathbf r\approx \mathbf t_i$ for different tail entity $t_i$, 
thus resulting in very similar embeddings for these different entities. 
To solve this problem, variants like 
TransH \cite{wang2014knowledge}, TransR \cite{lin2015learning}, 
TransD \cite{ji2015knowledge} are introduced to project embeddings of head/tail entity $\mathbf h, \mathbf t$ into various spaces. 
By maximizing the scoring function for all positive triplets, the distance between $\mathbf h + \mathbf r$ and $\mathbf t$ in corresponding space can be reduced. 


\subsubsection{Semantic matching model}
Another group of scoring functions operate without the assumption that $\mathbf h+\mathbf r\approx \mathbf t$. 
Instead, they use similarity to measure the plausibility of triplets $(h, r, t)$. 
RESCAL \cite{nickel2011three} is the most original model. The entity embeddings $\mathbf h, \mathbf t$ are also continuous vectors in $\mathbb R^d$. But for each relation, it is represented as a matrix which models the pairwise interaction between every dimension in entity embedding space $\mathbb R^d$. Namely, the scoring function of a triplet $(h, r, t)$ is defined as
$f(h, r, t) = \mathbf h^\top \mathbf M_r \mathbf t$,
where the relation is represented as a matrix $\mathbf M_r\in \mathbb R^{d\times d}$. 
This scoring function captures pairwise interactions between all components of $\mathbf h$ and $\mathbf t$, 
which needs $ O(d^2)$ parameters per relation. 

Some simple and effective variants of RESCAL are DistMult \cite{yang2014embedding}, HolE \cite{nickel2016holographic} and ComplEx \cite{trouillon2016complex}. 
DistMult simplifies RESCAL by restricting the interaction matrix $\mathbf M_r$ into a diagonal matrix, 
which can reduce the number of parameters per relation from $ O(d^2)$ to $ O(d)$. HolE and ComplEx improves DistMult by modeling asymmetric relations.


\section{Proposed Model}
\label{sec:proposed}

In this section,
we first describe our key observations in Section~\ref{sec:closer},
which are ignored by existing works but are the main motivations of our work.
The proposed method is described in Section~\ref{sec:NSCaching},
where we show how challenges in negative sampling are addressed by cache.
Finally,
we show an interesting connection between NSCaching and self-pace learning \cite{kumar2010self} in Section~\ref{sec:self-pace},
which further explains the good performance. 

\subsection{Closer Look at Distribution of Negative Triplets}
\label{sec:closer}

Recall that, in Equation~\eqref{eq:nega}, 
the negative triplet $(\bar{h}, r, \bar{t}) \not\in \mathcal{S}$ is formed 
by replacing either the head or tail entity of 
a positive triplet $(h, r, t)\in\mathcal{S}$ with any other entities in $\mathcal E$. 
Before introducing the proposed method,
we analyze the distribution of scores for $(\bar{h}, r, \bar{t}) \in \bar{\mathcal{S}}_{(h,r,t)}$.

\begin{figure}[ht]
	\centering
	\subfigure[Different epochs.]
	{\includegraphics[width=0.24\textwidth]{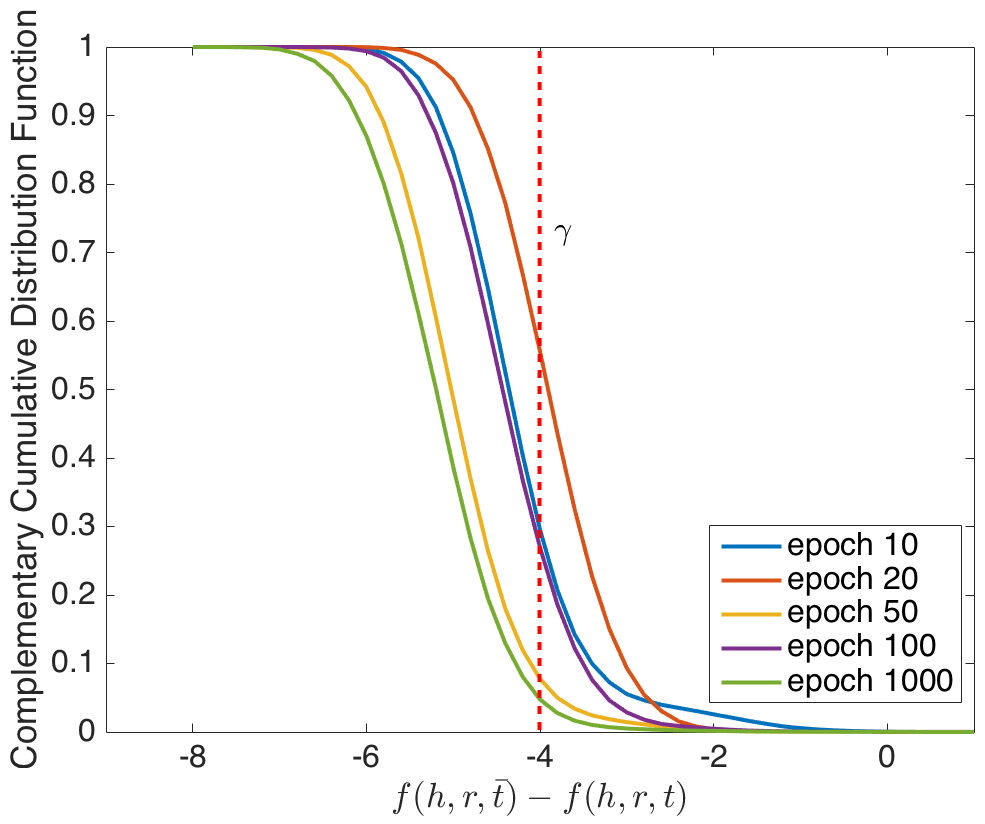}}
	\subfigure[Different triplets.]
	{\includegraphics[width=0.24\textwidth]{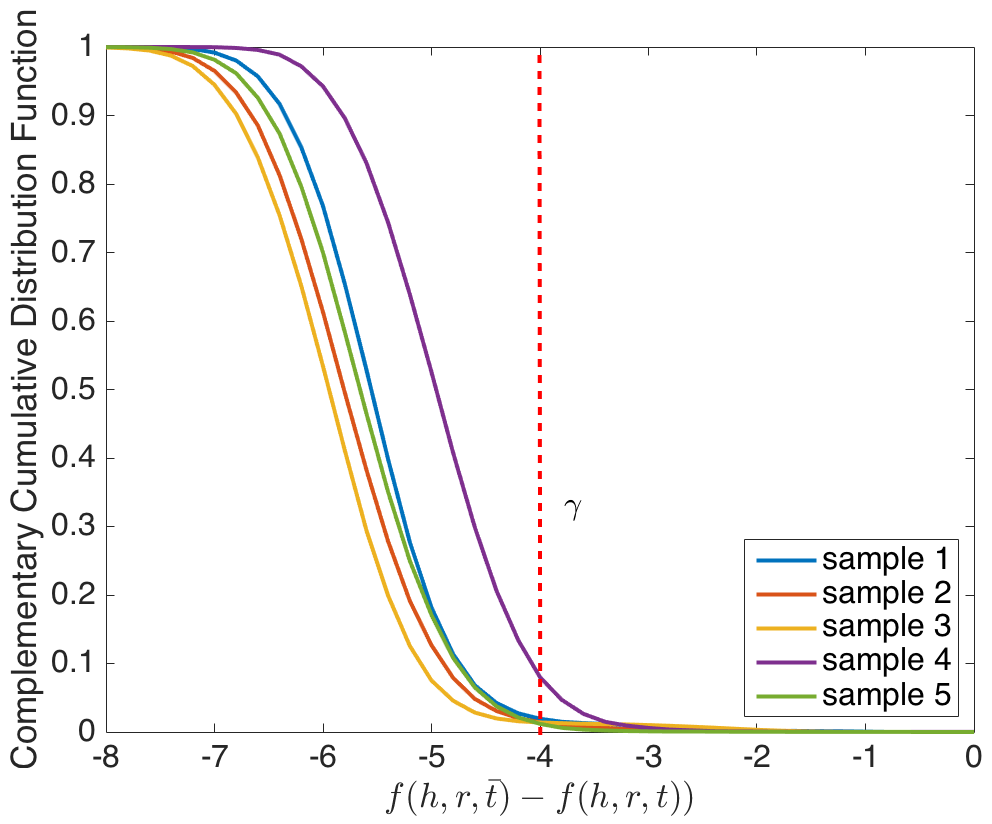}}
	
	
	\caption{Distribution of negative triplets on WN18 trained by Bernoulli-TransD
		(see Section~\ref{sssec:methods}). 
		For a given triplet $(h, r, t)$, we fix the head entity $h$ and relation $r$, and compute the distance $D_{(h,r,\bar{t})} = f(h, r, \bar{t}) - f(h, r, t)$.  
		We measure the \emph{complementary cumulative distribution function} (CCDF) $F_D(x) = P(D\geq x)$ to show the proportion of negative triplets that satisfy  $D\geq x$. 
		The red dashed line shows where the margin $\gamma$ lies. 
		(a) is the distribution of negative triplets in 6 timestamp of a certain triplet $(h, r, t)$. (b) is the negative sample distribution of 5 different triplets $(h, r, t)$ after the pretraining stage.}
	\label{fig-distance}
\end{figure}

Figure~\ref{fig-distance}(a) shows the changes in the distribution of negative samples
for one positive triplet;
and Figure~\ref{fig-distance}(b)
shows distributions of negative samples from different positive triplets.
Note that once the distance is larger than the margin $\gamma$,
i.e.,
the red vertical line,
the gradient of corresponding negative triplets will vanish to zero.
Indeed, we can see
the distribution changes during the training process; and
negative triplets with large scores are rare.
These observations are consistent with those ones in
\cite{wang2018incorporating,cai2018kbgan},
which further explain the vanishing gradient problem of uniform sampling,
as most sampled negative triplets will have small scores.

Although, 
the necessity of finding negative triplets with large scores from a dynamic distribution
is mentioned by above works,
they do not deeply study these distributions.
{\it
\noindent
\text{\bf Key Observations.}
The more important observations are:
\begin{itemize}
	\item The score distribution of negative triplets  is highly skew.
	
	\item Regardless of the training (Figure~\ref{fig-distance}(a)) and the choice of positive triplets (Figure~\ref{fig-distance}(b)),
	only a few negative triplets have large scores.
\end{itemize}}
Thus,
while GAN has strong ability to monitor the full generation process of negative triplets,
it wastes a lot of parameters and training time on learning
how negative triplets with small scores are distributed.
This is obviously not necessary. 
Besides,
reinforcement learning has been used once GAN is applied,
which increases the difficulties on training.
As a result,
\textit{is it possible to directly keep track of those negative triplets with large scores}?

\begin{table*}[ht]
	\caption{Comparison of the proposed approach with state-of-the-arts,
		which address the negative sample.
		Model parameters are based on TransE,
		$m$ is the size of mini-batch,
		$n$ is the epoch of lazy-update.
		}
	\centering
	\label{tab:compare}
	\renewcommand{\arraystretch}{1.3}
	\begin{tabular}{c | c | c | c | c | c }
		\hline
		                                  &             \multicolumn{2}{c|}{strategy}             &  \multicolumn{2}{c|}{minibatch computation}   & model                             \\ \hline
		                                  & negative sample & training                            & time                  & space                 & parameters                        \\ \hline
		            baseline              & uniform random  & gradient descent (from scratch)     & $O(m d)$              & $O(m d)$              & $ (|\mathcal E|+|\mathcal R|) d$  \\ \hline
		IGAN \cite{wang2018incorporating} & GAN             & reinforce learning (with pre-train) & $O(m |\mathcal E| d)$ & $O(m |\mathcal E| d)$ & $3(|\mathcal E|+|\mathcal R|) d $ \\ \hline
		    KBGAN \cite{cai2018kbgan}     & GAN             & reinforce learning (with pre-train) & $O(m N_1 d)$          & $O(m N_1  d)$         & $2(|\mathcal E|+|\mathcal R|) d $ \\ \hline
		            NSCaching              & using cache     & gradient descent (from scratch)     & $O( \frac{m}{n+1} (N_1+N_2)d)$      & $O(m (N_1+N_2)d)$    & $ (|\mathcal E|+|\mathcal R|) d$  \\ \hline
	\end{tabular}
\end{table*}

\subsection{NSCaching: the Proposed Method}
\label{sec:NSCaching}

In this section,
we describe the proposed method,
which addresses the aforementioned question.
The basic idea is very simple and intuitive.
Recall that the challenges in negative sampling 
are (i) how to model the dynamic distribution of negative triplets 
and (ii) how to sample negative triplets in an efficient way.
By considering the key observations,
we are motivated to use a small amount of memory,
which caches negative samples with large scores for each triplet in $\mathcal{S}$,
and sample the negative triplet directly from the cache. 
Algorithm \ref{alg-cache} shows the KG embedding framework based on our cache-based negative sampling scheme.
Note that the proposed sampling scheme does not depend on the choice of scoring functions,
all ones previously mentioned in Section~\ref{sec:scorefunc} can be used here.

\begin{algorithm}[ht]
	\caption{NSCaching: Cache-based KG embedding.}
	\label{alg-cache}
	\begin{algorithmic}[1]
		\REQUIRE training set $\mathcal{S} = \{(h, r, t)\}$, embedding dimension $d$, scoring function $f$, and head cache $\mathcal{H}$, tail cache $\mathcal{T}$.
		\STATE initialize embeddings for each $e \in \mathcal E$ and $r \in \mathcal R$,
		head-cache $\mathcal{H}$ and tail-cache $\mathcal{T}$;
		\FOR{$i = 1, \cdots, T$}
		\STATE sample a mini-batch $\mathcal S_{\text{batch}} \in \mathcal{S}$ of size $m$;
		\FOR{each $(h, r, t)\in\mathcal S_{\text{batch}}$}
		\STATE index the cache $\mathcal{H}$ by $(r, t)$ and $\mathcal{T}$ by $(h, r)$ to get the candidate sets of heads $\mathcal H_{(r,t)}$ and tails $\mathcal{T}_{(h,r)}$.
		\STATE \textit{core step}: sample $\bar{h}\in \mathcal H_{(r,t)}$ and $\bar{t}\in\mathcal{T}_{(h,r)}$.
		\STATE uniformly pick up $(\bar{h}, r, \bar{t}) \in \{(\bar{h}, r, t), (h, r, \bar{t})\}$.
		\STATE \textit{core step}: update the cache $\mathcal H_{(r,t)}$ and $\mathcal{T}_{(h,r)}$ using Algorithm~\ref{alg-cache-update};
		\STATE update embeddings 
		using Equation~\eqref{eq:graddist} or \eqref{eq:gradsema};
		\ENDFOR
		
		\ENDFOR
	\end{algorithmic}
\end{algorithm}

Basically,
as a negative triplet can be constructed by
either replacing the head or tail entity,
we maintain a head-cache $\mathcal{H}$ (indexed by $(r,t)$) and a tail-cache $\mathcal{T}$ (indexed by $(h,r)$),
which store $\bar{h} \in \mathcal{E}$ and $\bar{t} \in \mathcal{E}$ respectively. 
Each pair $(h, r)$ or $(r,t)$ corresponds to a unique index.
First,
when a positive triplet is received,
the corresponding cache containing candidates for negative triplets, 
i.e., 
$\mathcal H_{(r,t)}$ and $\mathcal{T}_{(h,r)}$,
are indexed in step~5.
A negative triplet is generated from $\mathcal H_{(r,t)}$ and $\mathcal{T}_{(h,r)}$ at step~6-7,
and then the cache is updated in step~8.
Finally,
the embeddings are updated based on the choice of scoring functions.

An overview of the proposed method
with state-of-the-arts are in Table~\ref{tab:compare}.
The main difference with general KG embedding framework in Algorithm~\ref{alg:basic} is step~5-8 in Algorithm~\ref{alg-cache},
where the sampling scheme is based on the cache instead.
Besides,
compared with 
previous complex GAN-based works \cite{wang2018incorporating,cai2018kbgan},
our method in Algorithm~\ref{alg-cache} acts like a discriminative and distilled model of GAN,
which only cares about negative triplets with large scores during the training.
Thus,
the proposed method,
i.e.,
NSCaching,
not only has fewer parameters,
but also can be easily trained from randomly initialized models (from the scratch).
Moreover,
experimental results in Section~\ref{sec:exp} show that NSCaching achieves the best performance.

However,
in order to achieve best performance,
we need to carefully design how to sample from the cache (step~6) and update the cache (step~8).
In the sequel,
we will describe the ``\textit{exploration and exploitation}'' inside these steps 
and how they are balanced in detail.
Then,
we give a time and space analysis of Algorithm \ref{alg-cache},
which further explain its efficiency and memory saving.
Note that,
we only discuss operations and designs for the head-cache $\mathcal{H}$ here,
as designs are the same for the tail-cache $\mathcal{T}$.

%

%
%
%
%

%
%

\subsubsection{Uniform sampling strategy from the cache (step~6)}
\label{sssec:sample}
Recall that only head $\bar{h}$ in negative triplets with large scores are in cache $\mathcal H_{(r,t)}$,
thus picking up any $\bar{h} \in \mathcal H_{(r,t)}$ probably avoids the vanishing gradient problem.
As larger scores also lead to bigger gradients,
a very natural scheme is to always sample the negative triplet with the largest score.

However,
as the distribution can change during the iterations of the algorithm,
the negative triplets in the cache may not be accurate enough for the sampling in the latest iteration.
Besides,
there are false negative triplets in the negative sample sets,
of which scores can also be very high \cite{wang2017knowledge}.
As a consequence, 
we also need to consider other triplets except the one with largest score in the cache.

\begin{algorithm}[ht]
	\caption{Updating head-cache (step~8).}
	\label{alg-cache-update}
	\begin{algorithmic}[1]
		\REQUIRE  head cache $\mathcal H_{(r, t)}$ of size $N_1$, triplet $(h, r, t)\in \mathcal S$.
		\STATE initialize $\tilde{\mathcal H}_{(r, t)} \leftarrow \emptyset$;
		\STATE uniformly sample a subset $\mathcal R_m\subset \mathcal E$ with $N_2$ entities;
		\STATE $\hat{\mathcal H}_{(r, t)} \leftarrow \mathcal H_{(r, t)} \cup \mathcal R_m$;
		\STATE compute the score $f(\bar{h}, r, t)$ for all $\bar{h} \in \hat{\mathcal H}_{(r,t)}$;
		\FOR{$i = 1, \cdots, N_1$}
			\STATE sample $\bar{h} \in \hat{\mathcal H}_{(r, t)}$ with probability in Equation~\eqref{eq:prupc};
			\STATE remove $\bar{h}$ from $\hat{\mathcal H}_{(r, t)}$;
			\STATE $\tilde{\mathcal H}_{(r, t)} \leftarrow \tilde{\mathcal H}_{(r, t)} \cup \bar{h}$;
		\ENDFOR
		\STATE update by ${\mathcal H}_{(r, t)} \leftarrow \tilde{\mathcal H}_{(r, t)}$.
	\end{algorithmic}
\end{algorithm}

This raises the question that
how to keep the balance between \textit{exploration} 
(i.e., explore all the possible high-quality negative samples)
and \textit{exploitation} (i.e., sample the largest score negative triplet in cache).

These motivate us to use uniformly random sampling scheme in step~6.
It is simple, efficient,
and does not introduce any bias into the selection process.
Indeed,
a stronger scheme can be sampling based on triplets' scores,
where larger score indicates higher probability to be sampled.
However,
it has extra memory costs as scores needs to be stored as well.
Moreover,
it introduces bias causing by dynamic changing distribution and false negative triplets,
which leads to inferior performance as shown in Section~\ref{sec:exp:sample-cache}.



\subsubsection{Importance sampling strategy to update the cache (step~8)}
\label{sssec:update}
As mentioned in Section~\ref{sec:uniframe},
the cache needs to be dynamically 
changed during the iterations of the algorithm.
Otherwise,
while negative triplets are kept in $\mathcal{H}_{(r,t)}$,
sampling from cache is still a scheme with fixed distribution,
which eventually suffers from vanishing gradient problem.
Thus,
we need to refresh the cache in each iteration.
Moreover,
the cache needs to be updated in an efficient way.

The proposed importance sampling (IS) strategy is presented in 
Algorithm~\ref{alg-cache-update}.
First,
we uniformly sample a subset $\mathcal R_m\subset \mathcal E$ of size $N_2$ (step~2),
then union it with $\mathcal{H}_{(r,t)}$ and obtain $\hat{\mathcal{H}}_{(r,t)}$.
The scores for all triplets in $\hat{\mathcal{H}}_{(r,t)}$ are evaluated in step~4.
After that,
we construct a subset $\tilde{\mathcal H}_{(r,t)}$ from $\hat{\mathcal{H}}_{(r,t)}$
by sampling entries in $\hat{\mathcal{H}}_{(r,t)}$ without replacement $N_1$ times following 
probability 
\begin{equation}
p(\bar{h}|(t, r)) = \frac{\exp(f(\bar{h}, r, t))}{\sum_{h_i\in\hat{\mathcal H}_{(r, t)} } \exp(f(\bar{h}_i, r, t))}.
\label{eq:prupc}
\end{equation}
Finally,
$\tilde{\mathcal H}_{(r,t)}$ is returned as the updated head-cache.


Note that exploration and exploitation also need to be carefully balanced in Algorithm~\ref{alg-cache-update}.
As the cache needs to be updated,
we have to sample from $\mathcal{E}$,
and uniform sampling is chosen due to its efficiency.
Thus,
a bigger $N_1$ implies more exploitation,
while a larger $N_2$ leads to more exploration.
In step~6,
indeed, 
uniform sampling or keeping triplets with top $N_1$ scores can be alternative choices.
However,  
both of them are inappropriate.
First,
uniformly sampling is obviously not proper,
as triplets in $\mathcal{H}_{(r,t)}$ have much larger scores than those in $\mathcal{R}_{m}$.
Then,
deterministically sampling top $N_1$ is not appropriate as well,
which again dues to the existence of false negative triplets (Section~\ref{sssec:sample}).
All above concerns will also be empirically studied in experiments Section~\ref{sec:exp}.


\subsubsection{Space and time complexities}
\label{sssec:complexity}
Here,
we analyze the space and time complexities of NSCaching (Algorithm~\ref{alg-cache}).
Comparing with basic Algorithm~\ref{alg:basic},
the main additional cost by introducing cache comes from Algorithm \ref{alg-cache-update} in step~8. 
In Algorithm \ref{alg-cache-update},
the time complexity of computing the score of $N_1+N_2$ candidate triplets $f(\bar{h}, r, t)$ is $O((N_1+N_2)d)$.
The cost of step~6 contains two parts,
i.e., normalization of the score and uniform sampling,
they take $O(N_1+N_2)$ and $O(N_1)$ respectively,
which are very small.
Thus, the total cost of introducing cache is $O((N_1+N_2)d)$ for one triplet.
We can lazily update the cache $n$ epochs later rather than immediately updating,
which can further reduce update complexity to $O((N_1+N_2)d/(n+1))$.

As for space complexity, 
evaluating the scores for $N_1+N_2$ candidate triplets takes $O((N_1+N_2)d)$ space. 
Since we only store indices in the cache, 
it takes $ O(|\mathcal S|N_1)$ space to store these indices for negative triplets.
However, since there are many one-to-many, many-to-one and many-to-many relations, 
the cost will be smaller than 
$ O(|\mathcal S|N_1)$ and the cache does not need to be stored in memory.
In our experiments,
values of $N_1$ and $N_2$ used on WN18 and FB15K are both $50$, 
which is much smaller than the number of entities.

In comparison, to generate one negative triplet, the generator in IGAN \cite{wang2018incorporating} costs $O(|\mathcal E|d)$ time since it needs to compute the distribution over all entities. 
KBGAN \cite{cai2018kbgan} needs $O(N_1 d)$ cost for measuring a candidate set of $N_1$ triplets. The additional space cost for IGAN and KBGAN is also $O(|\mathcal E|d)$ and $O(N_1 d)$ respectively.
Finally,
the comparisons are summarized in Table~\ref{tab:compare} with TransE as the scoring function. 

\subsubsection{Discussion on the Convergence}

{
Both the baseline KG embedding models \cite{wang2017knowledge} and NSCaching use stochastic gradient descent (SGD) for model training.
While there is no theoretical guarantee,
SGD has been applied on many nonconvex and complex models \cite{kingma2014adam},
where the convergence is empirically observed,
including the baseline KG embedding model \cite{bordes2013translating,bordes2014semantic,wang2014knowledge,fan2014transition,trouillon2016complex,nickel2016holographic,liu2017analogical}.
The only difference of NSCaching to that baseline model is how to sample negative triplets.

Besides,
since NSCaching samples negative triplets with larger scores,
its gradients have larger magnitude than that of baseline approach. 
This also prevents NSCaching from being early stopped by the sampling process 
and helps to converge with higher testing performance that of baseline models.
The above are all empirically shown and studied in Section~\ref{sec:exp}.

}

\subsection{Connection to Self-Pace Learning}
\label{sec:self-pace}

The main idea 
of self-paced (or curriculum) learning \cite{bengio2009curriculum,kumar2010self}
is to pick up easy samples first,
and then gradually switch to hard ones.
In this way,
the classifier can first identify the rough position where the decision boundary should locate,
and then the boundary can be further refined near hard examples.
It is very effective for complex and noncovex models.

Recently,
it is also introduced into network embedding 
and a big improvement on embedding's quality has been reported \cite{gao2018self}.
Besides,
GAN is also used to monitor the distribution of edges in the network,
and negative edges with scores above one threshold are sampled from the generator in GAN. 
Self-paced learning is achieved by increasing the threshold during the training of embedding \cite{gao2018self}.
Thus,
we can see neither KBGAN nor IGAN has benefited from self-paced learning.

In contrast,
our caching scheme can explicitly benefit from it.
{
The reason is that 
the embedding model only has weak discriminative ability in the beginning of the training. 
Thus,
while there are still a lot of negative triplets with large scores,
it is more likely that they are easy ones as most of negative samples are easy.
However, 
as training goes on,
those easy samples will gradually have small scores and are removed from the cache.
These mean NSCaching will learn from easy samples first,
but then gradually focus on hard ones,
which is exactly the principle of self-paced learning.
}
The above explanations are also verified by
experiments,
where we can see 
the negative triplets in the cache change from easy to hard ones (Section~\ref{sec:visself})
and NSCaching training from scratch can already achieve better performance
than IGAN and KBGAN with pre-training (Section~\ref{ssec:compstate}).

\section{Experiments}
\label{sec:exp}

In this section,
we carry empirical study of our method. 
All algorithms are written in Python with PyTorch framework \cite{paszke2017automatic} and run on a TITAN Xp GPU.

\subsection{Experiment Setup}
\label{secc:exp-setup}

\subsubsection{Datasets}
{
Four datasets are used here,
i.e., WN18, FB15K and their variants WN18RR, FB15K237.
WN18 and FB15K are firstly introduced in \cite{bordes2013translating}.
They are widely tested among the most famous Knowledge Graph embedding learning works
\cite{bordes2013translating,ji2015knowledge,trouillon2016complex,wang2018incorporating,cai2018kbgan}. 
WN18RR and FB15K237 are variants that remove near-duplicate or inverse-duplicate relations from WN18 and FB15K, and are introduced by \cite{wang2018evaluating} and  \cite{toutanova2015observed} respectively.
The two variants are harder and more realistic.
Their statistics are shown in Table \ref{tb-dataset}.
}

\begin{table}[ht]
	\centering
	\caption{Detailed information of the datasets used in experiments}
	\label{tb-dataset}
	\begin{tabular}{c|ccccc}
		\hline
		Dataset & \#entity & \#relation & \#train & \#valid & \#test \\ 
		\hline
		WN18 & 40,943 & 18 & 141,442 & 5,000 & 5,000 \\
		WN18RR & 93,003 & 11 & 86,835 & 3,034 & 3,134 \\
		FB15K & 14,951 & 1,345 & 484,142 & 50,000 & 59,071 \\
		FB15K237 & 14,541 & 237 & 272,115 & 17,535 & 20,466 \\
		\hline 
	\end{tabular}
\end{table}

Specifically,
WN18 and WN18RR are subsets of Wordnet \cite{miller1995wordnet}, which is a large lexical database of English. The entities correspond to word senses, and relations mean the lexical relation between them.
FB15K and FB15K237 are subsets of Freebase dataset \cite{bollacker2008freebase} which contains general facts of the world.
Freebase keeps growing until January 2014 and it now contains approximately 44 million topics and 2.4 billion triplets. 

\subsubsection{Tasks}
{Following previous KG embedding works \cite{bordes2013translating,wang2014knowledge,ji2015knowledge,trouillon2016complex}, 
	and the GAN-based works \cite{wang2018incorporating,cai2018kbgan}, 
	we test the performance on \emph{link prediction} task. 
	This is also the testbed to measure KG embedding models.}
Link prediction aims to predict the missing entity $h$ or $t$ for a positive triplet $(h, r, t)$. 
In this task, we measure the rank of head entity $h$ and tail entity $t$ among all the entity sets.
Thus, link prediction emphasizes the rank of the correct entity 
rather than their concrete scores.


\subsubsection{Performance measurements}
As in previous works \cite{bordes2013translating,trouillon2016complex,wang2018incorporating,cai2018kbgan}
, we evaluate different models based on the following metrics:
\begin{itemize}
\item Mean reciprocal ranking (MRR):
It is computed by average of the reciprocal ranks $1/|\mathcal S|\sum_{i=1}^{|\mathcal S|}\frac{1}{\text{rank}_i}$
where $\text{rank}_i, i\in \{ 1,\dots, |\mathcal S| \} $ is a set of ranking results;

\item Hit@10: 
It is the percentage of appearance in top-$k$: $1/|\mathcal S| \sum_{i=1}^{|\mathcal S|}\mathbb I(\text{rank}_i<k)$, where $\mathbb I(\cdot)$ is the indicator function;

\item Mean rank (MR): It is computed by $\frac{1}{|\mathcal S|}\sum_{i=1}^{|\mathcal S|}{\text{rank}_i}$.
Smaller value of MR tends to infer better results.
\end{itemize}

MRR and Hit@$10$ measure the top rankings of positive entity in different level. 
Hit@10 cares about general top rankings, and the top~1 samples contribute most to MRR.
The larger value of MRR and Hit@$10$ indicates better performance.
To avoid underestimating the performance of different models, 
we report the performance in a ``Filtered'' setting, 
i.e., all the corrupted triplets that exist in train, valid and test set are filtered out \cite{wang2018incorporating,cai2018kbgan}. 
Note that, 
MR is not a good metric, 
as it is easily influenced by false positive samples.
We report it here to keep consistency with existing literatures 
\cite{wang2018incorporating,cai2018kbgan}.

\subsubsection{Choices of the scoring function}
A large amount of scoring functions have been proposed in literature, including 
translational distance models TransE \cite{bordes2013translating}, TransH \cite{wang2014knowledge}, TransR \cite{lin2015learning}, TransD \cite{ji2015knowledge}, TranSparse \cite{ji2016knowledge}, TransM \cite{fan2014transition}, ManifoldE \cite{xiao2016from}, etc., 
and semantic matching models RESCAL \cite{nickel2011three}, DistMult \cite{yang2014embedding}, HolE \cite{nickel2016holographic}, ComplEx \cite{trouillon2016complex}, ANALOGY \cite{liu2017analogical}, etc. 
All these methods are summarized in a recent survey \cite{wang2017knowledge}.
Follow \cite{cai2018kbgan,wang2018incorporating},
in the sequel, 
TransE, TransH, TransD, DistMult and ComplEx will be used as scoring functions for comparison
(see their definitions in Table~\ref{tb-score-func}).


\begin{table}[ht]
	\centering
	\caption{Definitions of different scoring functions.
		All model parameter are real values,
		except ComplEx are complex values.
		$\text{Re}(\cdot)$ takes the real part out of complex numbers,
		$\text{diag}(\mathbf{r})$ constructs a diagonal matrix with $\mathbf{r}$.}
	\label{tb-score-func}
	\renewcommand{\arraystretch}{1.3}
	\begin{tabular}{c|c|c}
		\hline
		model     &           scoring function            &                                                                   definition                                                                   \\ \hline
		translational & TransE \cite{bordes2013translating} &                                               $\left\|\mathbf h+\mathbf r-\mathbf t \right\|_1$                                                \\ \cline{2-3}
		distance    &   TransH \cite{wang2014knowledge}   &       $\left\|\mathbf h \!- \! \mathbf w_r^\top \mathbf h\mathbf w_r\!+\!\mathbf r\!-\!(\mathbf t\!-\!\mathbf w_r^\top \mathbf t\mathbf w_r) \right\|_1$        \\ \cline{2-3}
		&    TransD \cite{ji2015knowledge}    & $\left\|\mathbf h\!+\!\mathbf w_r \mathbf w_h^\top\mathbf h \!+\!\mathbf r\!-\!(\mathbf t\!+\!\mathbf w_r \mathbf w_t^\top\mathbf t) \right\|_1$ \\ \hline
		semantic    &  DistMult \cite{yang2014embedding}  &                                                   $\mathbf{h} \cdot \text{diag}(\mathbf{r}) \cdot \mathbf{t}^{\top}$                                                    \\ \cline{2-3}
		matching    & ComplEx \cite{trouillon2016complex} &                                              $\text{Re}(\mathbf{h} \cdot \text{diag}(\mathbf{r}) \cdot \mathbf{t}^{\top})$                                              \\ \hline
	\end{tabular}
\end{table}



\subsection{Comparison with State-of-the-arts}
\label{ssec:compstate}

In this section,
we focus on the comparison with state-of-the-arts methods.
Hyper-parameters of \textit{NSCaching} are studied in Section~\ref{sec:ablation}.


\begin{table*}[ht]
\centering
\caption{Comparison of various algorithms on the four Dataset. 
	Performance of the \textit{predefined} model is included as reference.
	As code of IGAN is not available, 
	its performance is directly copied from \cite{wang2018incorporating}.
	Note that MRR, and those on WN18RR, FB15K237 datasets are not reported as they are not shown \cite{wang2018incorporating}.
	Bold number means the best performance, and underline means the second best.}
\label{tb-perf}
\renewcommand{\arraystretch}{1.3}
\begin{tabular}{c|c c|c c c|ccc|ccc|ccc}
	\hline
	         scoring          &     \multicolumn{2}{c|}{Dataset}     &                   \multicolumn{3}{c|}{WN18}                   &                  \multicolumn{3}{c|}{WN18RR}                   &                    \multicolumn{3}{c|}{FB15K}                    &                 \multicolumn{3}{c}{FB15K237}                  \\ \cline{2-15}
	        functions         &     \multicolumn{2}{c|}{Metrics}     &        MRR         & MR              & $\!\!\!$Hit@10$\!\!\!$ &        MRR         &        MR        & $\!\!\!$Hit@10$\!\!\!$ &        MRR         &        MR         & $\!\!\!$ Hit@10$\!\!\!$ &        MRR         &       MR        & $\!\!\!$Hit@10$\!\!\!$ \\ \hline
	 \multirow{8}{*}{TransE}  &   \multicolumn{2}{c|}{pretrained}    &       0.4213       & \textbf{217}    & 91.50                  &       0.1753       & \underline{4038} &         44.48          &       0.4679       &    \textbf{60}    &          74.70          &       0.2262       &       237       &         38.64          \\
	                          &    \multicolumn{2}{c|}{Bernoulli}    &       0.5001       & 249             & 94.13                  &       0.1784       &  \textbf{3924}   &         45.09          &       0.4951       &       {65}        &         {77.37}         &       0.2556       &       197       &         41.89          \\
	                          &      KBGAN      & $\!\!\!\!$pretrain &       0.6880       & 293             & \textbf{94.92}         &       0.1864       &       4420       &         45.39          &       0.4858       &        82         &          77.02          &       0.2938       &       628       &         46.69          \\
	                          &                 & $\!\!\!\!$scratch  &       0.6606       & 301             & 94.80                  &       0.1808       &       5356       &         43.24          &       0.3771       &        335        &          72.67          &       0.2926       &       722       &         46.59          \\
	                          & $\!\!$NSCaching & $\!\!\!\!$pretrain &  \textbf{0.7867}   & 271             & \textbf{66.62}         &  \textbf{0.2048}   &       4404       &   \underline{47.38}    &  \textbf{0.6475}   &  \underline{62}   &     \textbf{81.54}      &  \textbf{0.3004}   & \underline{188} &   \underline{47.36}    \\
	                          &                 & $\!\!\!\!$scratch  & \underline{0.7818} & 249             & 94.63                  & \underline{0.2002} &       4472       &     \textbf{47.83}     & \underline{0.6391} &  \underline{62}   &    \underline{80.95}    & \underline{0.2993} &  \textbf{186}   &     \textbf{47.64}     \\
	                          &      IGAN       & $\!\!\!\!$pretrain &       ------       & \underline{240} & 91.3                   &       ------       &      ------      &         ------         &       ------       &        81         &          74.0           &       ------       &     ------      &         ------         \\
	                          &                 & $\!\!\!\!$scratch  &       ------       & 244             & 92.7                   &       ------       &      ------      &         ------         &       ------       &        90         &          73.1           &       ------       &     ------      &         ------         \\ \hline
	 \multirow{8}{*}{TransH}  &   \multicolumn{2}{c|}{pretrained}    &       0.4527       & \textbf{233}    & 92.71                  &       0.1755       &       5646       &         43.30          &       0.4316       &       {58}        &          73.98          &       0.2222       &       223       &         38.80          \\
	                          &    \multicolumn{2}{c|}{Bernoulli}    &       0.5206       & 288             & 94.52                  &       0.1862       &  \textbf{4113}   &         45.09          &       0.4518       &       {60}        &          76.55          &       0.2329       &       202       &         40.10          \\
	                          &      KBGAN      & $\!\!\!\!$pretrain &       0.6168       & 335             & 94.84                  &       0.1923       &       4708       &         45.31          &       0.4262       &        86         &          75.91          &       0.2807       &       401       &        {46.39}         \\
	                          &                 & $\!\!\!\!$scratch  &       0.6018       & 288             & 94.60                  &       0.1869       &       4881       &         44.81          &       0.3364       &        311        &          72.53          &       0.2779       &       455       &         46.19          \\
	                          & $\!\!$NSCaching & $\!\!\!\!$pretrain &  \textbf{0.8063}   & 286             & \textbf{95.32}         & \underline{0.2038} & \underline{4425} &     \textbf{48.04}     & {\textbf{0.6520}}  &   \textbf{{54}}   &     \textbf{81.96}      & \underline{0.2812} & \underline{187} &   \underline{46.48}    \\
	                          &                 & $\!\!\!\!$scratch  & \underline{0.8038} & 266             & \underline{95.29}      &  \textbf{0.2041}   &       4491       &     \textbf{48.04}     & \underline{0.6391} &    \textbf{54}    &    \underline{81.05}    &  \textbf{0.2832}   &  \textbf{185}   &     \textbf{46.59}     \\
	                          &      IGAN       & $\!\!\!\!$pretrain &       ------       & \underline{258} & 94.0                   &       ------       &      ------      &         ------         &       ------       &        81         &          77.0           &       ------       &     ------      &         ------         \\
	                          &                 & $\!\!\!\!$scratch  &       ------       & 276             & 86.9                   &       ------       &      ------      &         ------         &       ------       &        90         &          73.3           &       ------       &     ------      &         ------         \\ \hline
	 \multirow{8}{*}{TransD}  &   \multicolumn{2}{c|}{pretrained}    &       0.4426       & \underline{243} & 92.69                  &       0.1782       &       4955       &         42.18          &       0.4320       &       {59}        &          73.98          &       0.2244       &       215       &         39.53          \\
	                          &    \multicolumn{2}{c|}{Bernoulli}    &       0.5093       & 256             & 94.61                  &       0.1901       &       3555       &         46.41          &       0.4529       &       {63}        &          76.55          &       0.2451       & \underline{188} &         42.89          \\
	                          &      KBGAN      & $\!\!\!\!$pretrain &       0.6130       & 307             & 94.92                  &       0.1917       &       3785       &         46.49          &       0.4069       &        75         &          74.27          &       0.2487       &       798       &         44.33          \\
	                          &                 & $\!\!\!\!$scratch  &       0.5950       & 332             & 94.68                  &       0.1875       &       4083       &         46.41          &       0.3151       &        184        &          69.77          &       0.2465       &       825       &         44.40          \\
	                          & $\!\!$NSCaching & $\!\!\!\!$pretrain &  \textbf{0.8022}   & 295             & \underline{94.99}      &  \textbf{0.2013}   &  \textbf{2952}   &   \underline{48.36}    &  \textbf{0.6567}   &    \textbf{54}    &     \textbf{82.02}      &  \textbf{0.2883}   &  \textbf{184}   &     \textbf{48.33}     \\
	                          &                 & $\!\!\!\!$scratch  & \underline{0.7994} & 286             & \textbf{95.16}         &  \textbf{0.2013}   & \underline{3104} &     \textbf{48.39}     & \underline{0.6415} &  \underline{58}   &    \underline{81.32}    & \underline{0.2863} &       189       &   \underline{47.85}    \\
	                          &      IGAN       & $\!\!\!\!$pretrain &       ------       & 248             & 93.3                   &       ------       &      ------      &         ------         &       ------       &        79         &         {77.6}          &       ------       &     ------      &         ------         \\
	                          &                 & $\!\!\!\!$scratch  &       ------       & \textbf{221}    & 93.0                   &       ------       &      ------      &         ------         &       ------       &        89         &          74.0           &       ------       &     ------      &         ------         \\ \hline\hline
	\multirow{6}{*}{DistMult} &   \multicolumn{2}{c|}{pretrained}    &       0.6340       & 1174            & 92.28                  &       0.3765       &  \textbf{7405}   &         44.85          &       0.5004       &        176        &          77.46          &       0.2247       &       408       &         36.03          \\
	                          &    \multicolumn{2}{c|}{Bernoulli}    &       0.7918       & \textbf{862}    & \textbf{93.38}         &       0.3964       & \underline{7420} &         45.25          &       0.5698       &        148        &         {76.32}         &       0.2491       &       280       &         42.03          \\
	                          &      KBGAN      & $\!\!\!\!$pretrain &       0.6955       & 1143            & 93.11                  &       0.3849       &       7586       &         44.32          &       0.5568       &        201        &          75.57          &       0.2670       &       370       &         45.34          \\
	                          &                 & $\!\!\!\!$scratch  &       0.7275       & \underline{794} & 93.08                  &       0.2039       &      11351       &         29.52          &       0.4227       &        321        &          64.35          &       0.2272       &      {276}      &         39.91          \\
	                          & $\!\!$NSCaching & $\!\!\!\!$pretrain & \underline{0.8297} & 1038            & \underline{93.83}      &  \textbf{0.4148}   &       7477       &     \textbf{45.80}     & \underline{0.7177} &    \textbf{98}    &     \textbf{84.56}      &  \textbf{0.2882}   &  \textbf{265}   &     \textbf{45.79}     \\
	                          &                 & $\!\!\!\!$scratch  &  \textbf{0.8306}   & 827             & 93.74                  & \underline{0.4128} &       7708       &   \underline{45.45}    & {\textbf{0.7501}}  & \underline{{132}} &    \underline{84.36}    & \underline{0.2834} & \underline{273} &   \underline{45.56}    \\ \hline
	\multirow{6}{*}{ComplEx}  &   \multicolumn{2}{c|}{pretrained}    &       0.8046       & 1106            & 93.75                  &       0.3934       &       8259       &         41.63          &       0.5558       &        115        &          79.95          &       0.2201       &       418       &         35.55          \\
	                          &    \multicolumn{2}{c|}{Bernoulli}    &       0.9115       & \textbf{808}    & \textbf{94.39}         &       0.4431       &  \textbf{4693}   &     \textbf{51.77}     &       0.6713       &       {78}        &         {85.05}         &       0.2596       &       238       &         43.54          \\
	                          &      KBGAN      & $\!\!\!\!$pretrain &       0.8976       & 1060            & 93.73                  &       0.4287       &       6929       &         47.03          &       0.6254       &        162        &          80.95          &       0.2818       &       268       &         45.37          \\
	                          &                 & $\!\!\!\!$scratch  &       0.7233       & \underline{966} & 85.81                  &       0.3180       &       7528       &         35.51          &       0.5002       &        294        &          76.10          &       0.1910       &       881       &         32.07          \\
	                          & $\!\!$NSCaching & $\!\!\!\!$pretrain & \underline{0.9286} & 1079            & \underline{94.03}      &  \textbf{0.4487}   & \underline{4861} &   \underline{51.76}    & \underline{0.7459} &  \underline{123}  &    \underline{84.17}    & \underline{0.3017} &  \textbf{220}   &   \underline{47.75}    \\
	                          &                 & $\!\!\!\!$scratch  &  \textbf{0.9355}   & 1072            & 93.98                  & \underline{0.4463} &       5365       &         50.89          &  \textbf{0.7721}   &    \textbf{82}    &     \textbf{86.82}      &  \textbf{0.3021}   & \underline{221} &     \textbf{48.05}     \\ \hline
\end{tabular}
\end{table*}

\subsubsection{Compared methods}
\label{sssec:methods}

Following methods for negative sampling are compared:
\begin{itemize}
\item \textit{Bernoulli} \cite{wang2014knowledge}:
As a basic extension of the uniform sampling scheme used in TransE, 
Bernoulli sampling aims at reducing false negative labels by replacing the head or tail with different probability for one-to-many, many-to-one and many-to-many relations. 
Specifically, it samples $(\bar{h}, r, t)$ or $(h, r, \bar{t})$ under a predefined Bernoulli distribution. 
Since it is shown to be better than uniform sampling, we choose it as the basic random sampling scheme;

\item \textit{KBGAN} \cite{cai2018kbgan}:
This model firstly samples a set $\mathcal Neg$ uniformly from the whole entity set $\mathcal E$. 
Then head or tail entity is replaced with the entities in  $\mathcal Neg$ to form a set of candidate $(\bar{h}, r, t)$  and $(h, r, \bar{t})$. 
The generator in KBGAN tries to pick up one triplet among them. As proposed in \cite{cai2018kbgan}, we choose the simplest model TransE as the generator. For fair comparison, the size of set $\mathcal Neg$ is same as our cache size $N_1$.
We use the published code \footnote{\url{https://github.com/cai-lw/KBGAN}} and change the configure same as ours for fair comparison;

\item \textit{NSCaching} (Algorithm~\ref{alg-cache}):
As in Section \ref{sec:proposed}, 
the negative samples are formed by replacing the head entity $h$ or tail entity $t$ with one uniformly sampled from
head cache $\mathcal{H}$ or tail cache $\mathcal{T}$.
The cache is updated as in Algorithm \ref{alg-cache-update}. 
Note that we can also lazily update the cache several iterations later, which can further save time. 
However, we just report the result of immediate update, which is shown to be both effective and efficient.
We use $N_1=N_2=50$ and lazy-update with $n=0$ unless otherwise specified.
\end{itemize}

\begin{figure*}[ht]
	\centering
	\subfigure[WN18]
	{\includegraphics[width = 0.236\textwidth]{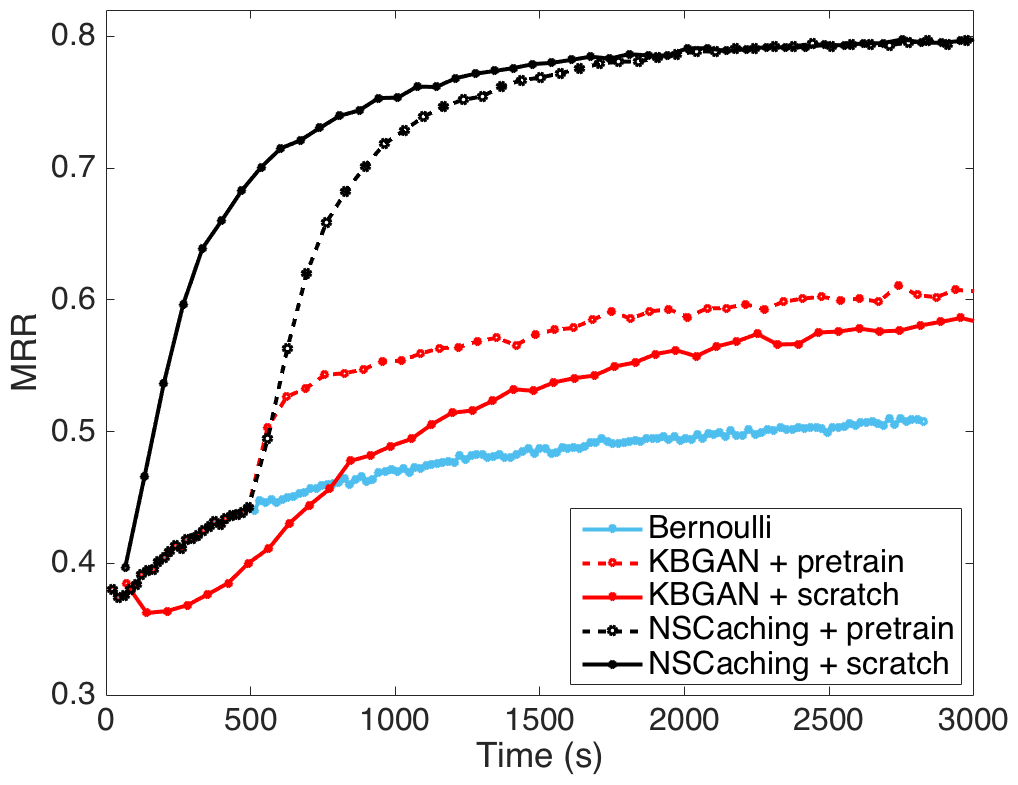}}
	\subfigure[WN18RR]
	{\includegraphics[width = 0.245\textwidth]{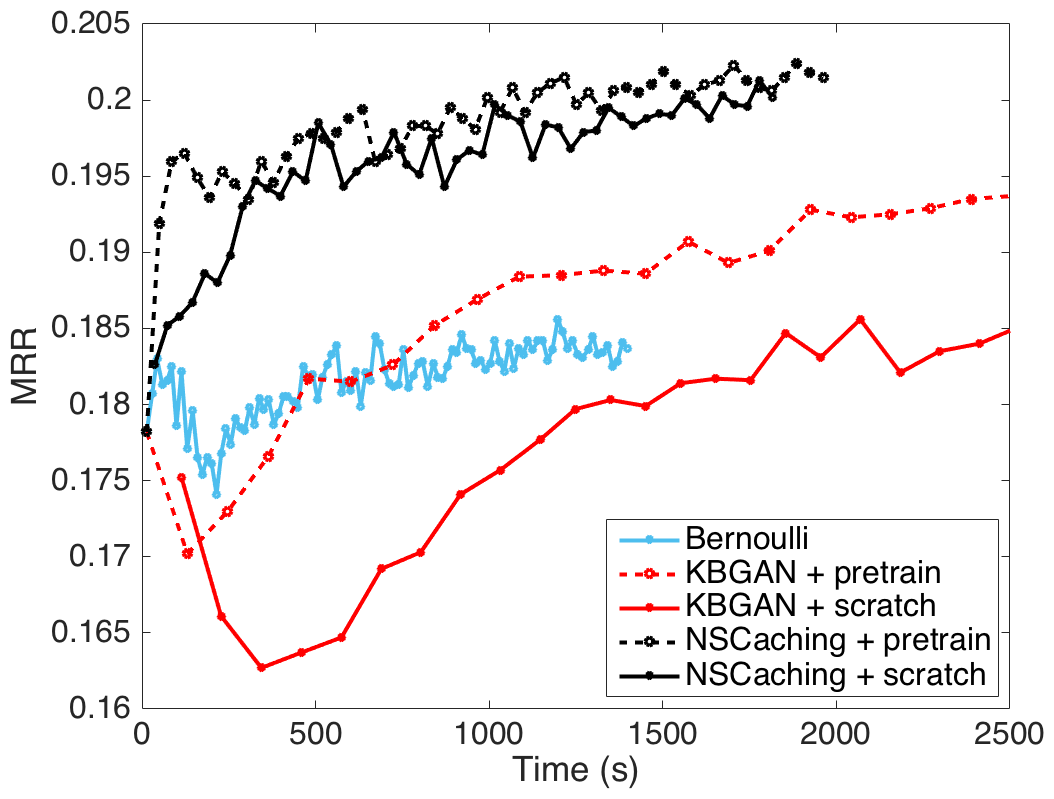}}
	\subfigure[FB15K]
	{\includegraphics[width = 0.235\textwidth]{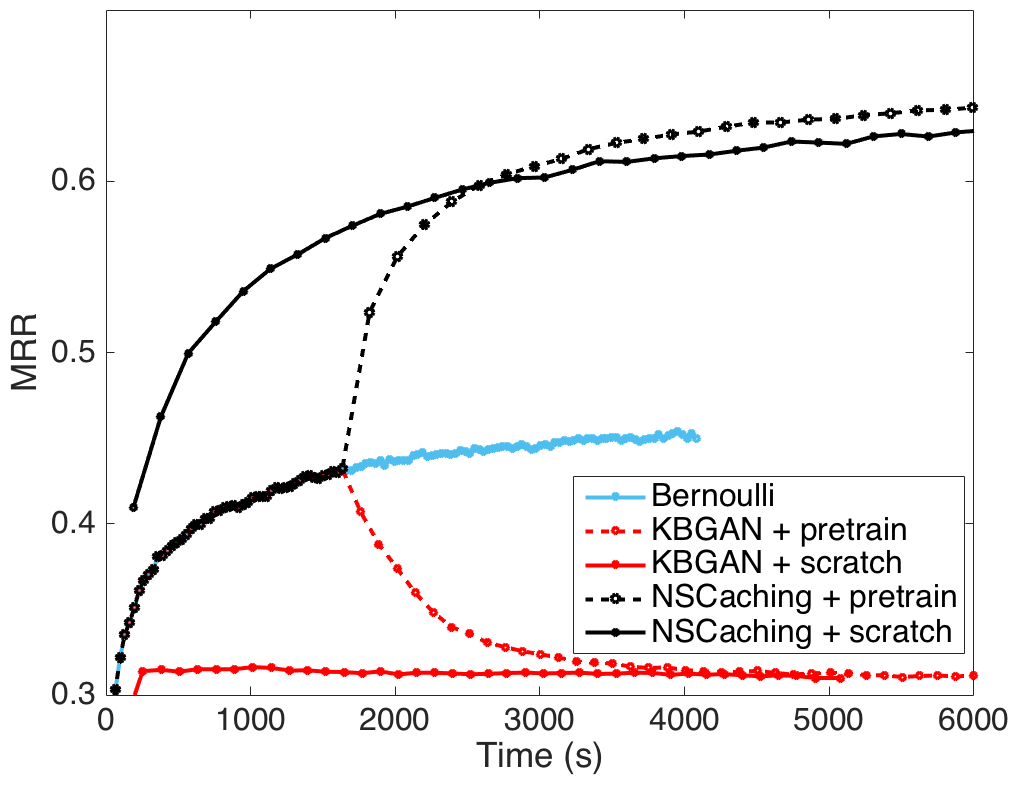}}
	\subfigure[FB15k237]
	{\includegraphics[width = 0.238\textwidth]{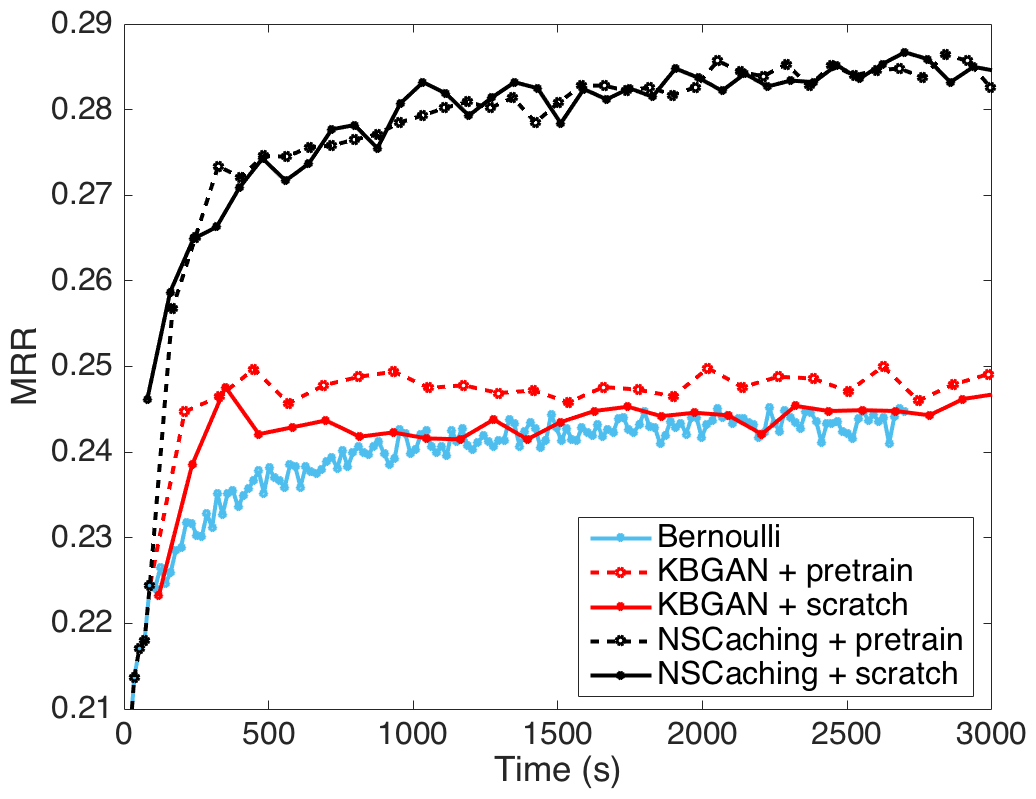}}
	
	\caption{Testing MRR performance v.s. clock time (in seconds) based on TransD.}
	\label{fig-perf-transd-mrr}
\end{figure*}

\begin{figure*}[ht]
	\centering
	\subfigure[WN18]
	{\includegraphics[width = 0.236\textwidth]{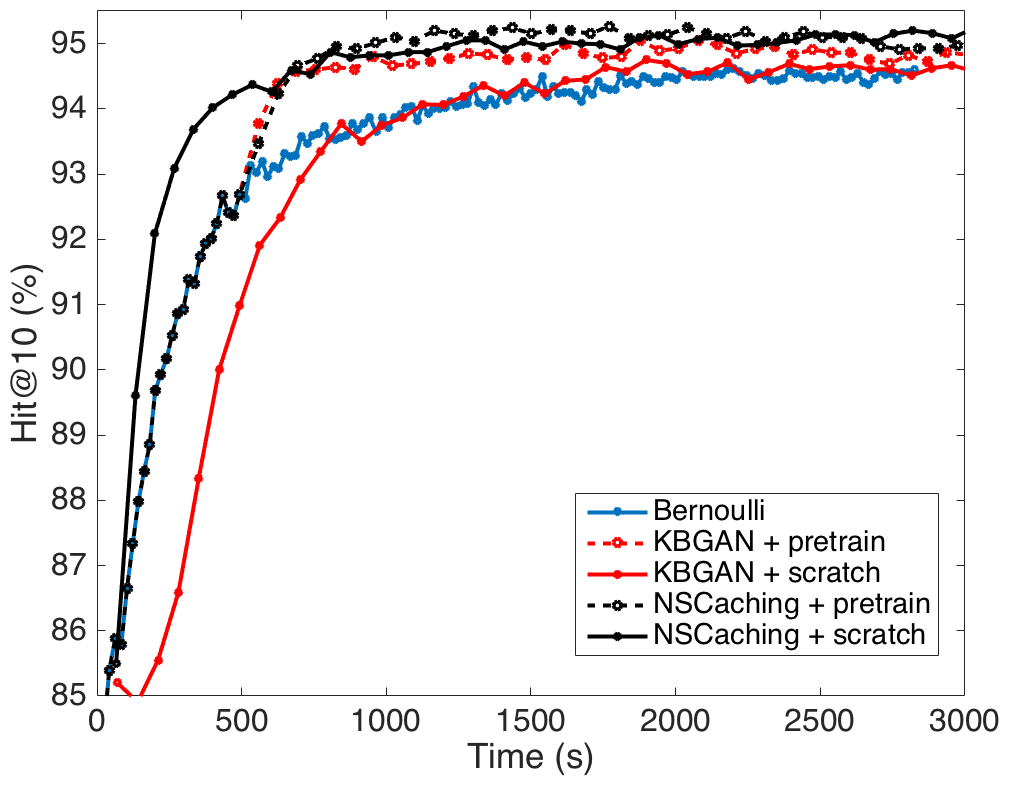}}
	\subfigure[WN18RR]
	{\includegraphics[width = 0.235\textwidth]{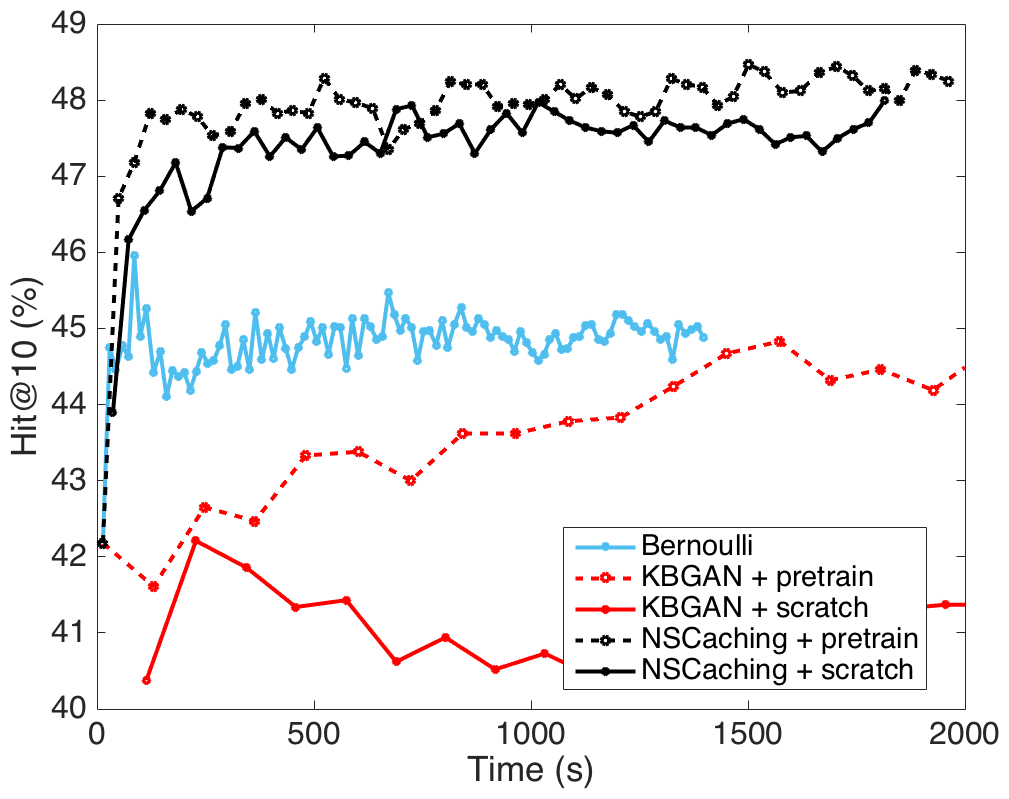}}
	\subfigure[FB15K]
	{\includegraphics[width = 0.235\textwidth]{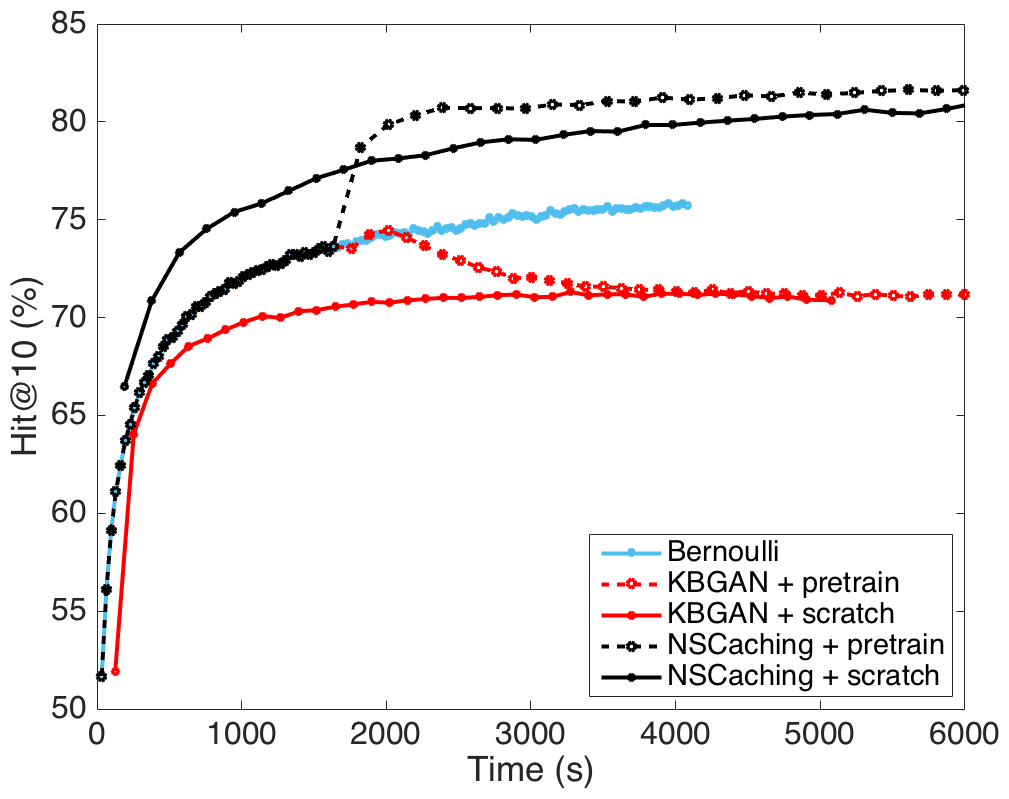}}
	\subfigure[FB15k237]
	{\includegraphics[width = 0.238\textwidth]{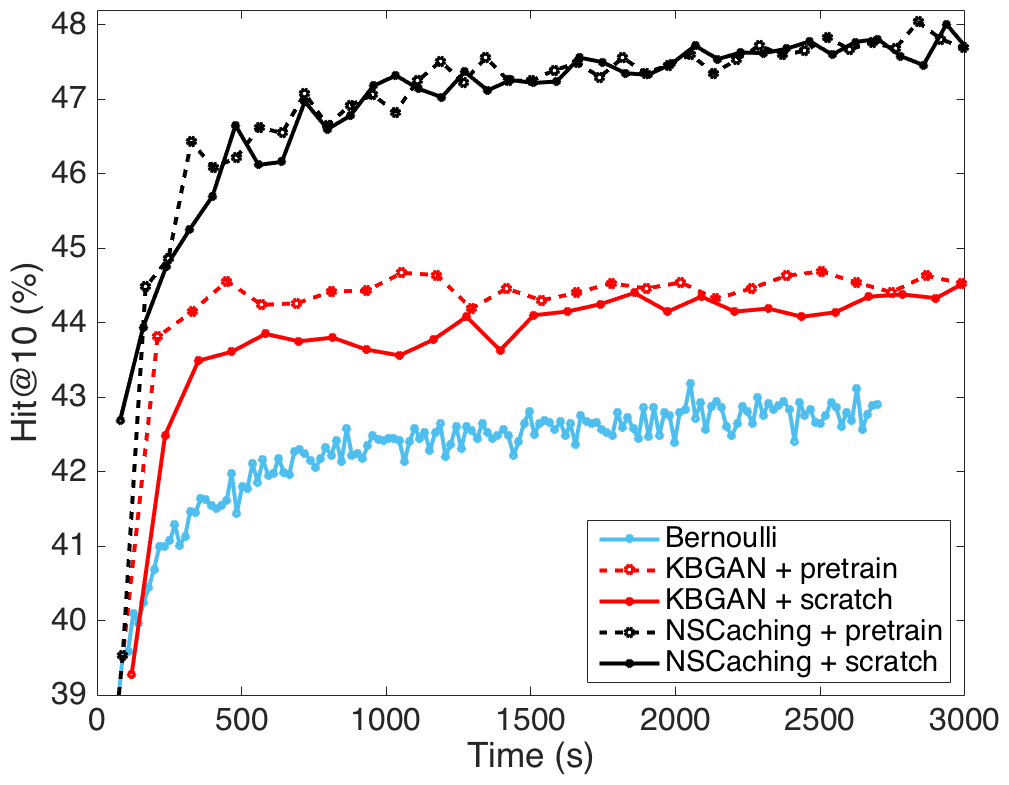}}
	
	\caption{Testing Hit@10 performance v.s. clock time (in seconds) based on TransD.}
	\label{fig-perf-transd-hit10}
\end{figure*}

\begin{figure*}[ht]
	\centering
	\subfigure[WN18]
	{\includegraphics[width = 0.236\textwidth]{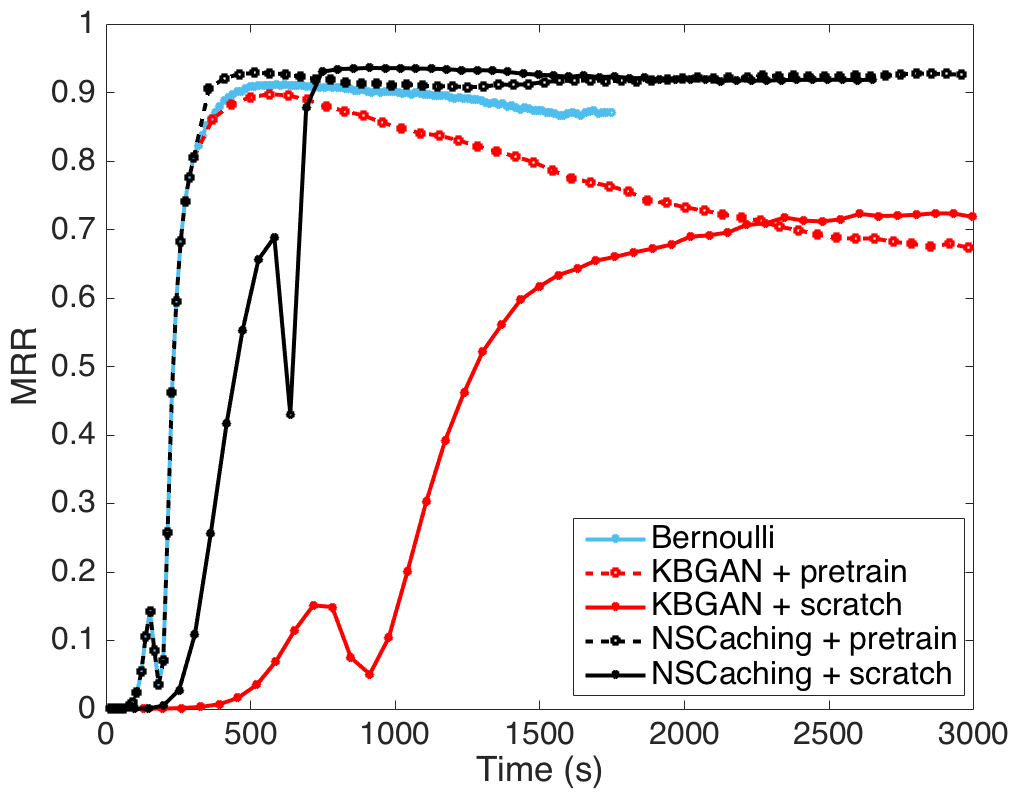}}
	\subfigure[WN18RR]
	{\includegraphics[width = 0.238\textwidth]{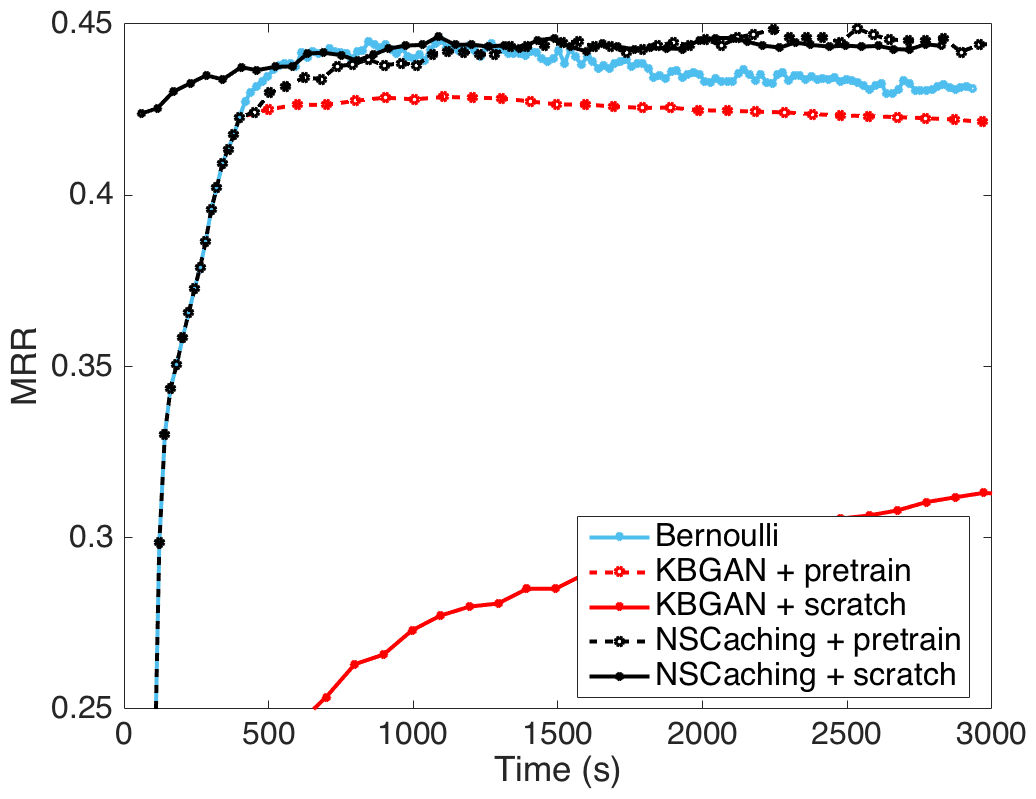}}
	\subfigure[FB15K]
	{\includegraphics[width = 0.235\textwidth]{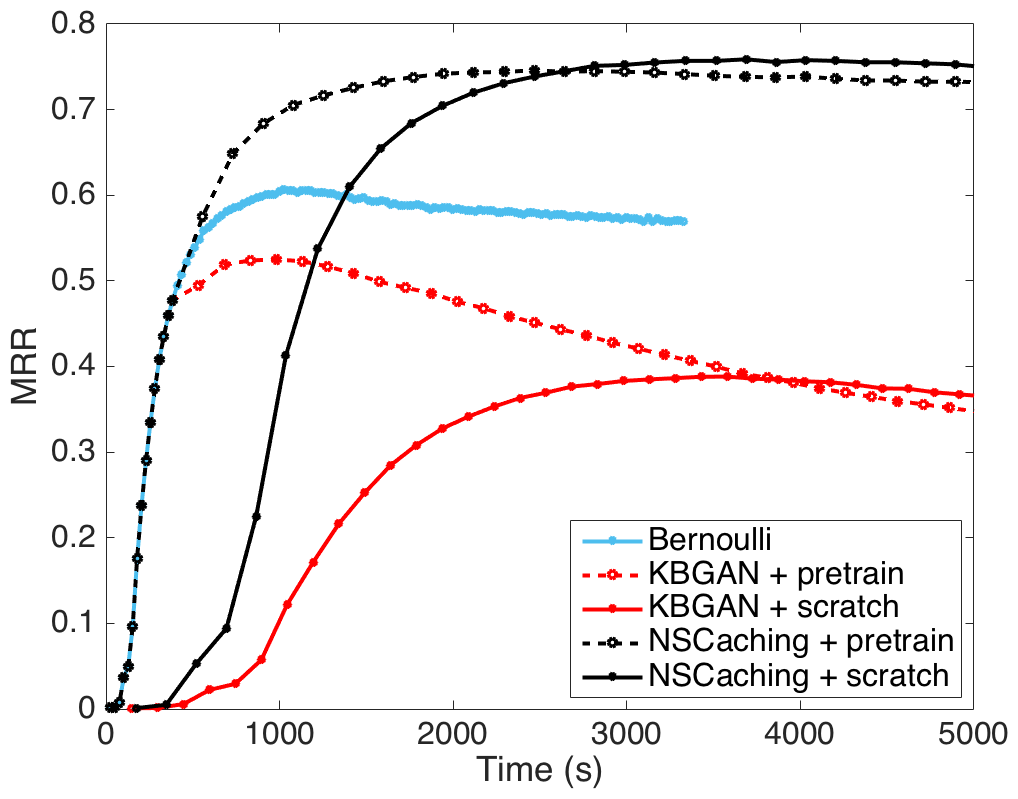}}
	\subfigure[FB15k237]
	{\includegraphics[width = 0.238\textwidth]{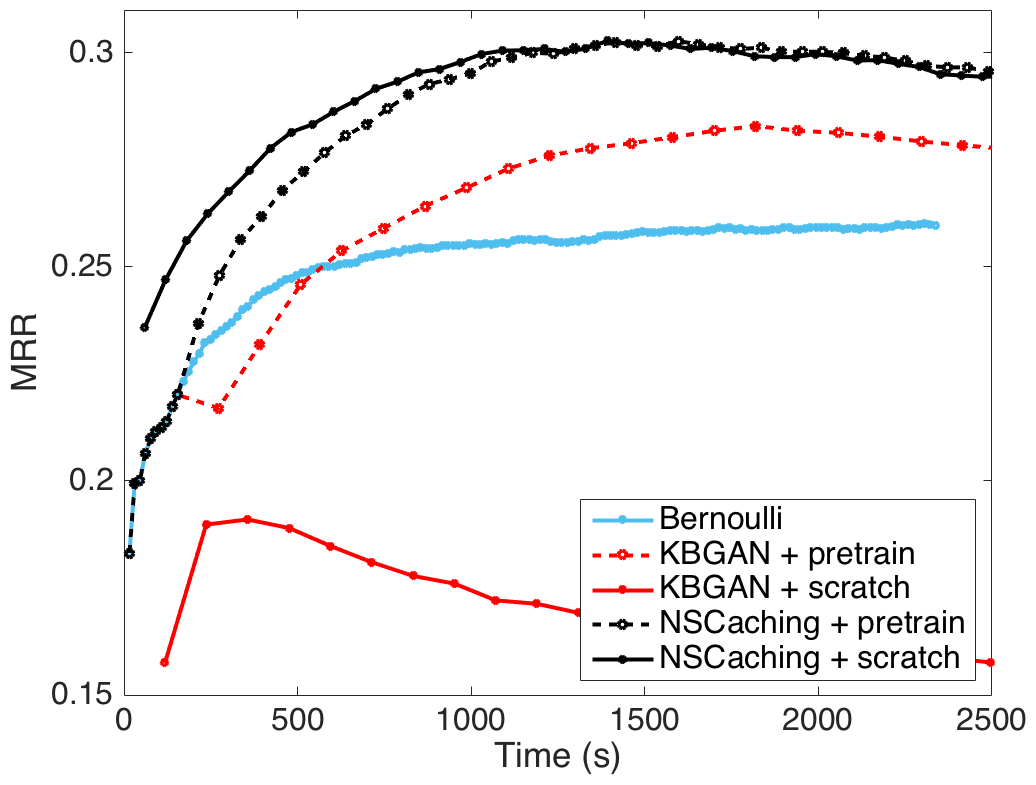}}
	
	\caption{Testing MRR performance v.s. clock time (in seconds) based on ComplEx.}
	\label{fig-perf-complex-mrr}
\end{figure*}

\begin{figure*}[ht]
	\centering
	\subfigure[WN18]
	{\includegraphics[width = 0.236\textwidth]{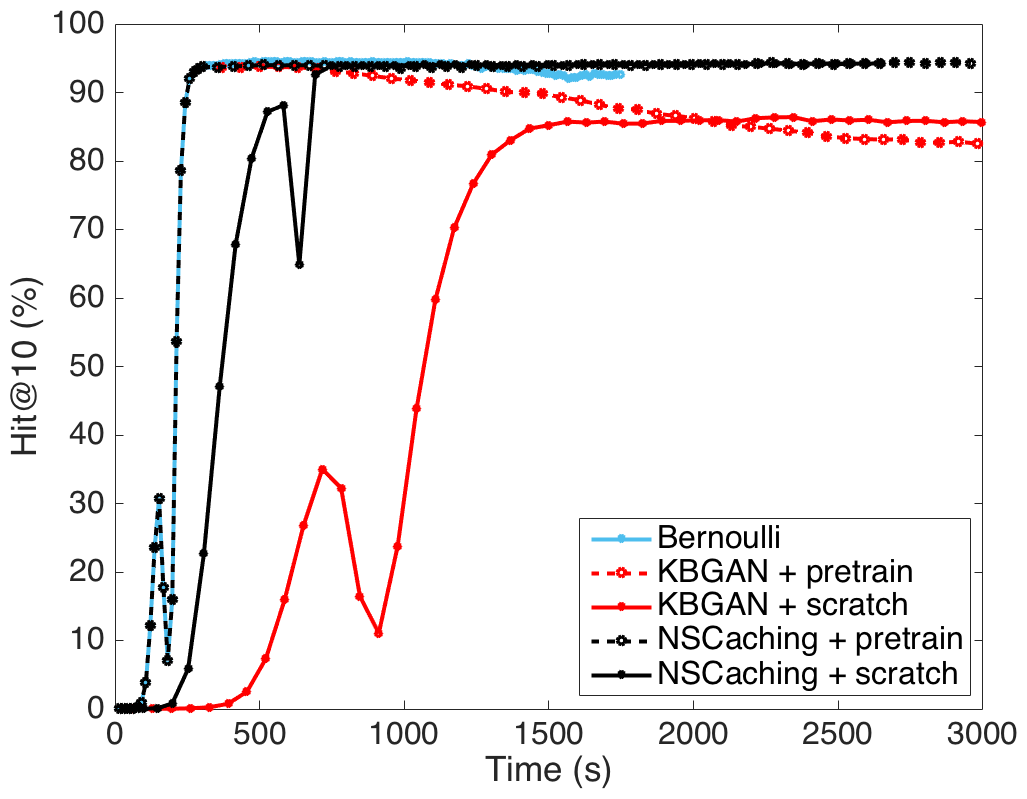}}
	\subfigure[WN18RR]
	{\includegraphics[width = 0.234\textwidth]{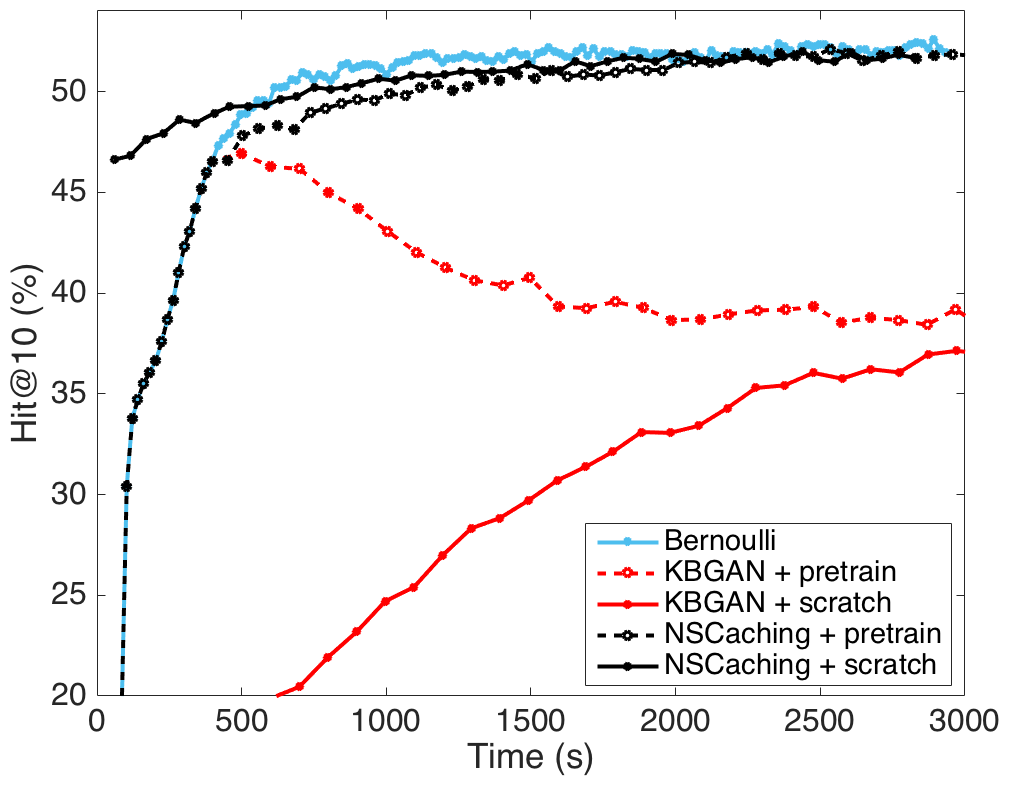}}
	\subfigure[FB15K]
	{\includegraphics[width = 0.235\textwidth]{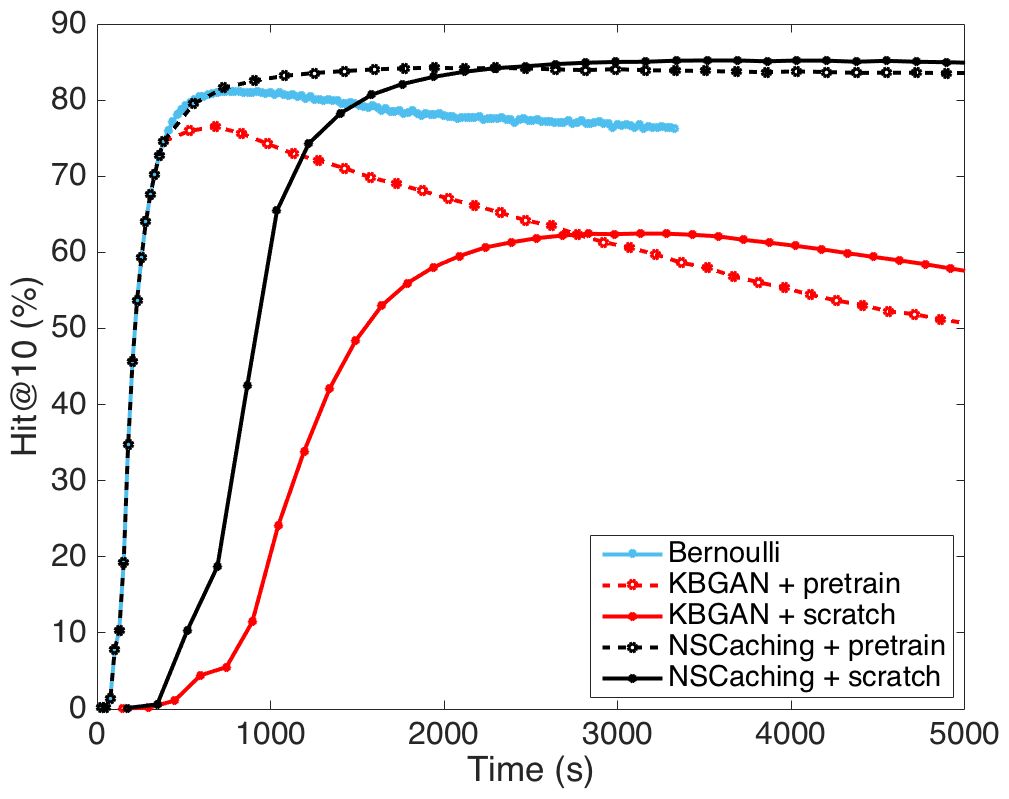}}
	\subfigure[FB15k237]
	{\includegraphics[width = 0.235\textwidth]{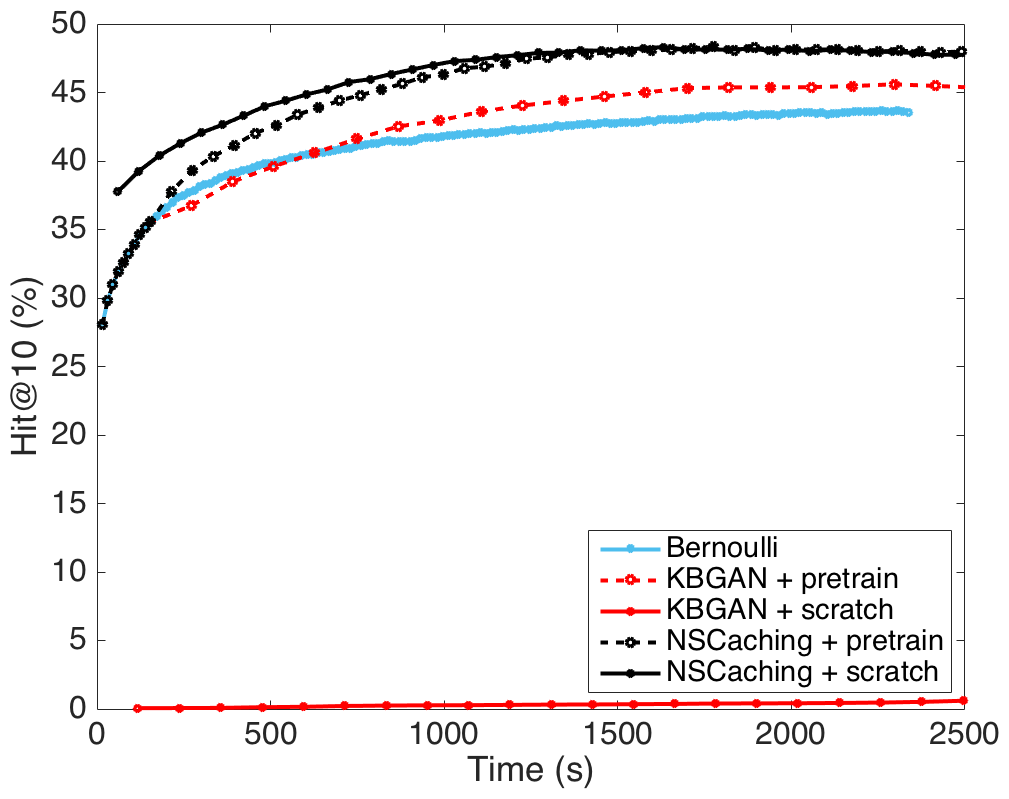}}
	
	\caption{Testing Hit@10 performance v.s. clock time (in seconds) based on ComplEx.}
	\label{fig-perf-complex-hit10}
\end{figure*}

As the source code of \textit{IGAN} \cite{wang2018incorporating} is not available, 
we do not compare with it here.
Instead,
we directly use the reported performance in the sequel.
Finally,
we also use Bernoulli sampling to choose between $(\bar{h}, r, t)$ and $(h, r, \bar{t})$ for \textit{KBGAN} and \textit{NSCaching}.

Besides,
as suggested in \cite{cai2018kbgan,wang2018incorporating},
two training strategies are used for \textit{KBGAN} and \textit{NSCaching},
i.e.,
\begin{itemize}
\item From \textit{scratch}: The embedding of relations and entities are initialized by the Xavier uniform initializer \cite{glorot2010understanding},
and the models
 (denoted as \textit{KBGAN + scratch} and \textit{NSCaching + scratch})
are directly applied to train the given KG; 

%
%
\item With \textit{pretrain}:
Same as \cite{cai2018kbgan,wang2018incorporating},
we firstly pretrain each scoring function under the baseline model, i.e. \textit{Bernoulli} sampling, 
several epochs on both data sets. 
We denote it as \textit{pretrained}. 
Then the obtained parameters are used to warm-start the given KG rather than from scratch.
We keep training based on the warm-started KG embedding 
and evaluate the performance under different models,
i.e., \textit{Bernoulli}, \textit{KBGAN + pretrain} and \textit{NSCaching + pretrain}.
Besides, the generator in \textit{KBGAN} is warm-started with corresponding TransE model.
\end{itemize}

\subsubsection{Hyper-parameter settings}

We use grid search to select the following hyper-parameters:
hidden dimension $d\in \{20, 50, 100, 200\}$,
learning rate $\eta \in$ $\{0.0001, 0.001, 0.01, 0.1\}$.
For translational distance models, we tune the margin value $\gamma \in \{1, 2, 3, 4\}$.
And for semantic matching models, we tune the penalty value $\lambda\in$ $\{0.001, 0.01, 0.1 \}$ \cite{trouillon2016complex}.
{We use Adam \cite{kingma2014adam}, 
	which is a popular variant of SGD algorithm for the training,
	and adopt its default settings, except for the learning rate.}
The best hyper-parameter is tuned under Bernoulli sampling scheme and evaluated by the MRR metric on validation set. 
We keep them fixed for the baseline methods \textit{Bernoulli}, \textit{KBGAN} and our proposed \textit{NSCaching}.
Following \cite{cai2018kbgan}, we save and record 
the \textit{pretrained} model after several initial training epochs.
	Then, \textit{Bernoulli} method keeps training until 3000 epochs; and 
	the results of \textit{KBGAN} and \textit{NSCaching} algorithm are evaluated within 1000 epochs,
	either from scratch or with pretrain.
All the recorded results are tested based on the best parameters chosen by the MRR value on validation set.

%
%
%
%

{
\subsubsection{Results on translational distance models}
The performance on link prediction is compared in Table~\ref{tb-perf}.
First, we can see that,
for the translational distance models (TransE, TransH, TransD), 
\textit{KBGAN}, \textit{NSCaching} and \textit{IGAN} (both \textit{with pretrain} and \textit{from scratch}) gain significant improvement upon the baseline scheme \textit{Bernoulli}, 
especially for the gaining on MRR, 
which is mainly influenced by top rankings.
This verifies the needs of using high-quality negative triplets during negative sampling and these methods can effectively pick up these negative triplets.

Then, \textit{IGAN} and \textit{KBGAN} with pretrain can perform better,
indicated by MRR and Hit@10,
than from scratch.
This shows pretrain is helpful for GAN based models. 
In comparison, the proposed \textit{NSCaching} trained from either state (pretrain or scratch)
can outperform \textit{IGAN} and \textit{KBGAN}.
Finally, we find that MR is not an appropriate metric, 
as many of the \textit{pretrained} models, which is not converged yet, 
show even smaller MR than the \textit{Bernoulli}.

Convergence of testing performance for various algorithms are shown in Figure~\ref{fig-perf-transd-mrr} and \ref{fig-perf-transd-hit10}.
We use TransD as it offers the best performance among the three translational distance models.
As can be seen, 
{
	all algorithms will converge to a stable testing MRR and Hit@10, which verifies the empirical convergence of Adam optimizer.
}
Then, while pretrain is a must for \textit{KBGAN} to achieve good performance,
\textit{NSCaching} can obtain good performance either from scratch or using pretrain.
Finally,
in all cases,
\textit{NSCaching} converges much faster and is more stable than both \textit{Bernoulli} and \textit{KBGAN}.

\subsubsection{Results on semantic matching models}
The performance is shown in the bottom rows of Table~\ref{tb-perf}. 
Same as the performance on translational distance models,
\textit{NSCaching} outperforms baseline scheme \textit{Bernoulli} significantly,
as indicated by the bold and underline numbers. 
However, \textit{KBGAN} does not show consistent performance.
It performs even worse than the \textit{Bernoulli} sampling scheme on WN18, WN18RR and FB15K,
\textit{KBGAN from scratch} even performs much worse than \textit{with pretrian}.
This observation further verifies the fact that GAN based methods usually suffer from instability and degeneracy.
This method needs careful balance between the generator and the target KG embedding model.
However, \textit{NSCaching} works consistently and performs the best among various settings.

Convergence of testing performance for various algorithms are shown in Figure~\ref{fig-perf-complex-mrr} and \ref{fig-perf-complex-hit10}.
We use ComplEx as the representative since it is much better than DistMult.
As can be seen,
both \textit{Bernoulli} and the proposed \textit{NSCaching} will converge to a stable state.
In the contrast, 
\textit{KBGAN} will turn down and overfit after several epochs.
However,
\textit{NSCaching}, either with pretrain or from scratch, leads the performance and is well adopted on the semantic matching models without further tuning.
}

\subsubsection{Results on triplets classification}

To further verify the quality of learned embedding, 
we test the learned embeddings on triplet classification task on WN18RR and FB15K237 datasets. 
This task is to confirm whether a given triplet $(h, r, t)$ is correct or not, 
i.e., binary classification on triplet \cite{wang2014knowledge}. 
In practice, it can help us to quickly answer the truth-or-false questions.
The WN18RR 
\footnote{\url{https://github.com/thunlp/OpenKE/blob/master/benchmarks/WN18RR/valid_neg.txt}}
and FB15K237 
\footnote{\url{https://github.com/thunlp/OpenKE/blob/master/benchmarks/FB15K237/valid_neg.txt}}
dataset released a set of positive and negative triplets,
which can be used to evaluate the performance on the classification task.
The decision rule of classification is as follows: for each $(h, r, t)$, if its score is no less than the relation-specific threshold $\sigma_r$, then predict positive. 
Otherwise, negative. The threshold $\sigma_r$ is determined according to 
maximizing the classification accuracy on the validation set.
As shown in Table~\ref{tb:classification}, NSCaching still outperforms the baselines. 
The new experiment further justifies that our proposed NSCaching can help learn a better embedding of the KG.

\begin{table}[H]
	\centering
	\caption{Comparison of various algorithms on tasks of relation prediction and triplet classification.} 
	\label{tb:classification}
	\scalebox{0.9}
	{
		\renewcommand{\arraystretch}{1.2}
		\begin{tabular}{c|cc|c|c}
			\hline
			scoring     function     &  \multicolumn{2}{c|}{Dataset}  &                {WN18RR}                   & {FB15K237}  \\ \hline
			\multirow{5}{*}{TransD}  & \multicolumn{2}{c|}{Bernoulli} &       86.81          &  78.24   \\
			&   KBGAN   &     pretrained     &  85.93    &  79.03   \\
			&           &      scratch       &   86.01   &79.05   \\
			& NSCaching &     pretrained     &          \textbf{87.84}          &   80.63   \\
			&           &      scratch       &      87.64          &  \textbf{80.69}   \\ \hline
			\multirow{5}{*}{ComplEx} & \multicolumn{2}{c|}{Bernoulli} &        84.48          & 77.64   \\
			&   KBGAN   &     pretrained     &  79.87  &    74.11  \\
			&           &      scratch       &    71.73  &   72.61    \\
			& NSCaching &     pretrained     &         \textbf{84.96}          &   79.88   \\
			&           &      scratch       &        84.83          &  \textbf{80.21}   \\ \hline
		\end{tabular}
	}
\end{table}

\subsection{Cache Update and Sampling Scheme}
\label{sec:ablation}

In Section~\ref{ssec:compstate},
we have shown that NSCaching achieves the best performance on four benchmark datasets.
Here,
we analyze design concerns on ``exploration and exploitation''
at step~6 and 8 in Algorithm~\ref{alg-cache}.
TransD and WN18 are used here.

\subsubsection{Uniform sampling from the cache (step~6)}
\label{sec:exp:sample-cache}

\begin{figure}[ht]
\subfigure[Diff. sampling strategies]
{\includegraphics[width=0.235\textwidth]{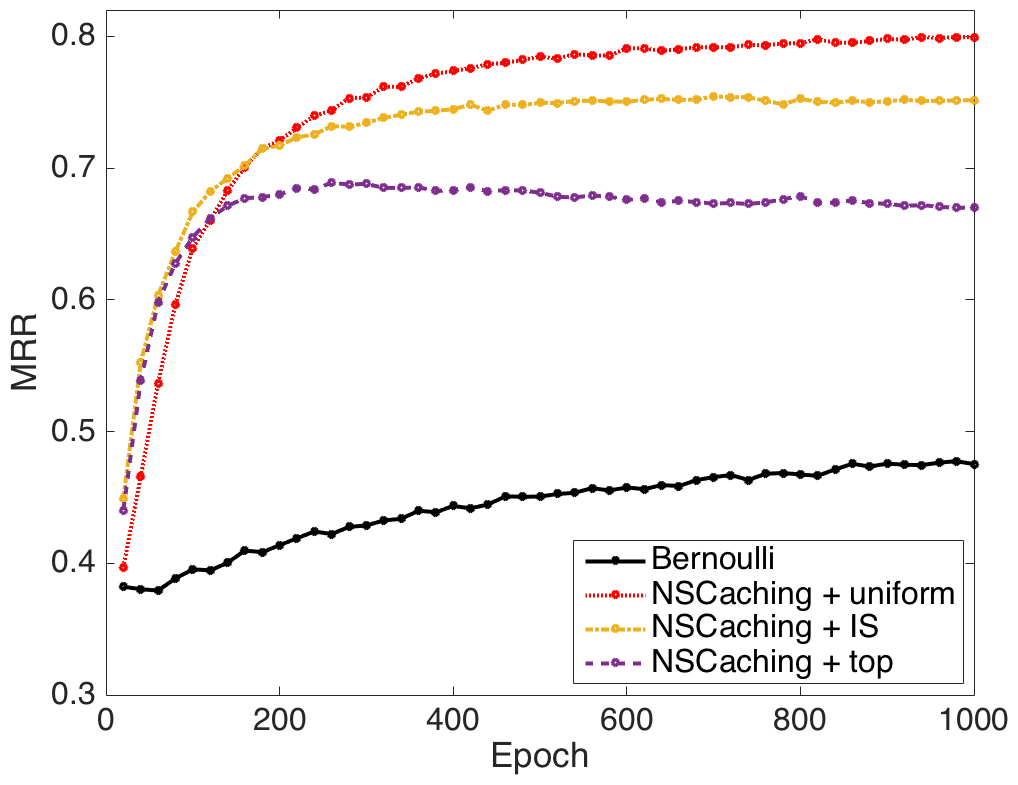} }
\subfigure[Diff. cache updating strategies]
{\includegraphics[width=0.234\textwidth]{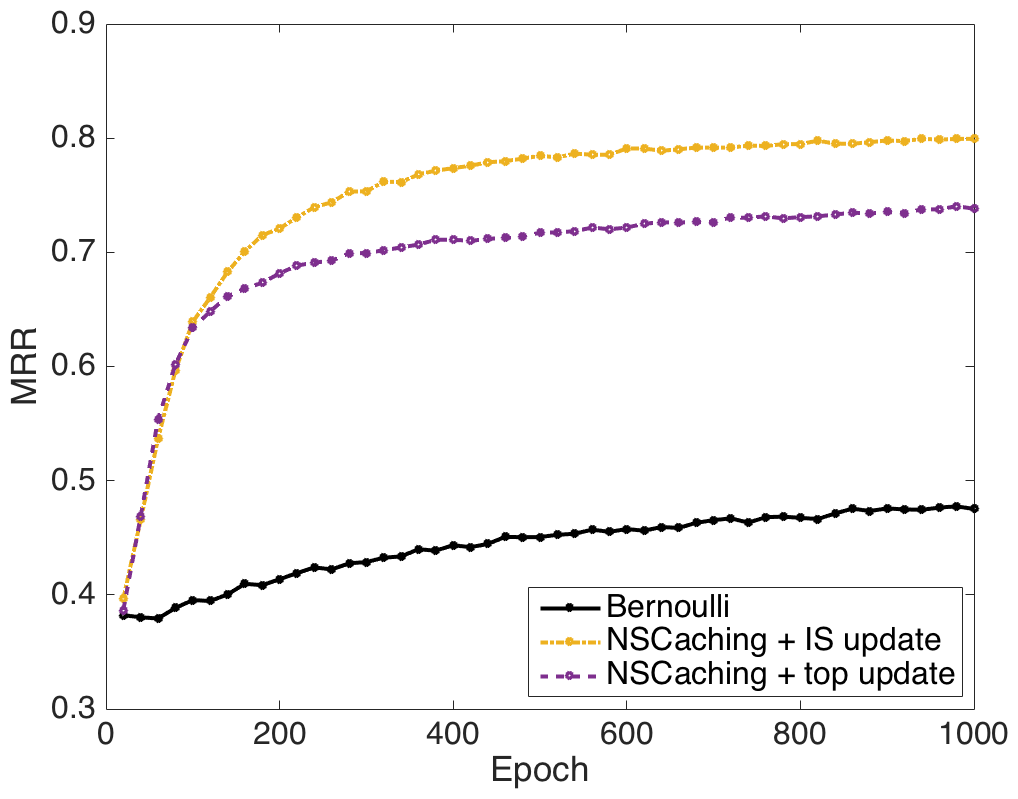} }
\caption{Comparison on testing MRR v.s. epoch of different sampling strategies and cache updating strategies. Evaluated by TransD model on WN18.}
\label{fig:ablation:sample}
\end{figure}

Given a cache, which stores high-quality negative samples, 
how to sample from it is the first question we care about. 
Recall that we discussed three strategies in Section~\ref{sssec:sample}, i.e., 
(i) uniform sampling from the cache (dented as ``uniform sampling''); 
(ii) importance sampling according to the score of each sample in cache (denoted as ``IS sampling''); and
(iii) top sampling, by choosing the sample with largest score (denoted as ``top sampling'').
Testing performance of MRR on WN18 trained by TransD are compared in Figure~\ref{fig:ablation:sample}.(a).
As can be seen,
top sampling has the worst performance,
and uniform sampling is the best.

To show how exploration and exploitation are balanced here,
we further compute two criterion to show the difference between these strategies. 
(i) Repeat ratio (denoted as ``RR''), 
which measures the percentage of repeated negative triplets $(\bar{h}, r, \bar{t})$ within $20$ epochs; and 
(ii) non-zero loss ratio (denoted as ``NZL''), which is the percentage of non-zero losses in same range. 
The value of RR is related to \emph{exploration}, if the number of repeated negative triplets is high, 
the negative samples only explore a small part of the sample spaces, thus result in worse exploration. 
NZL ratio measures \emph{exploitation}, a larger NZL means higher quality of picked negative samples.

\begin{figure}[ht]
\subfigure[RR v.s. epoch.]
{\includegraphics[width=0.225\textwidth]{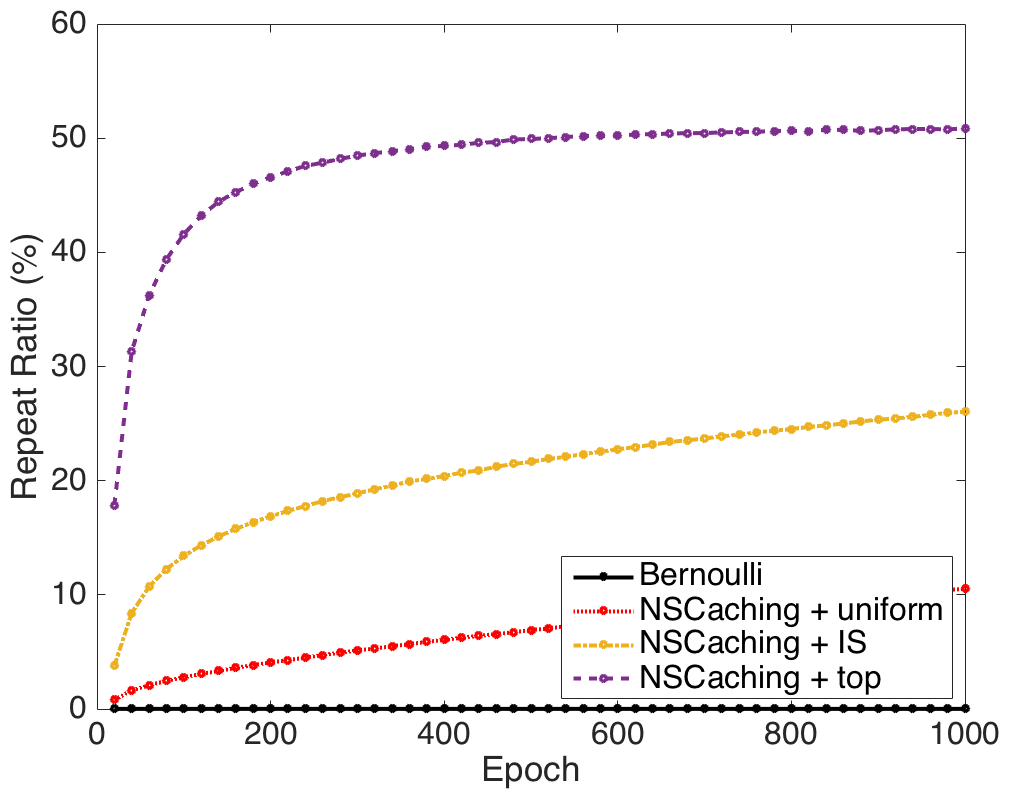}
\label{fig:sampleee:repeat}}
\subfigure[NZL v.s. epoch.]
{\includegraphics[width=0.225\textwidth]{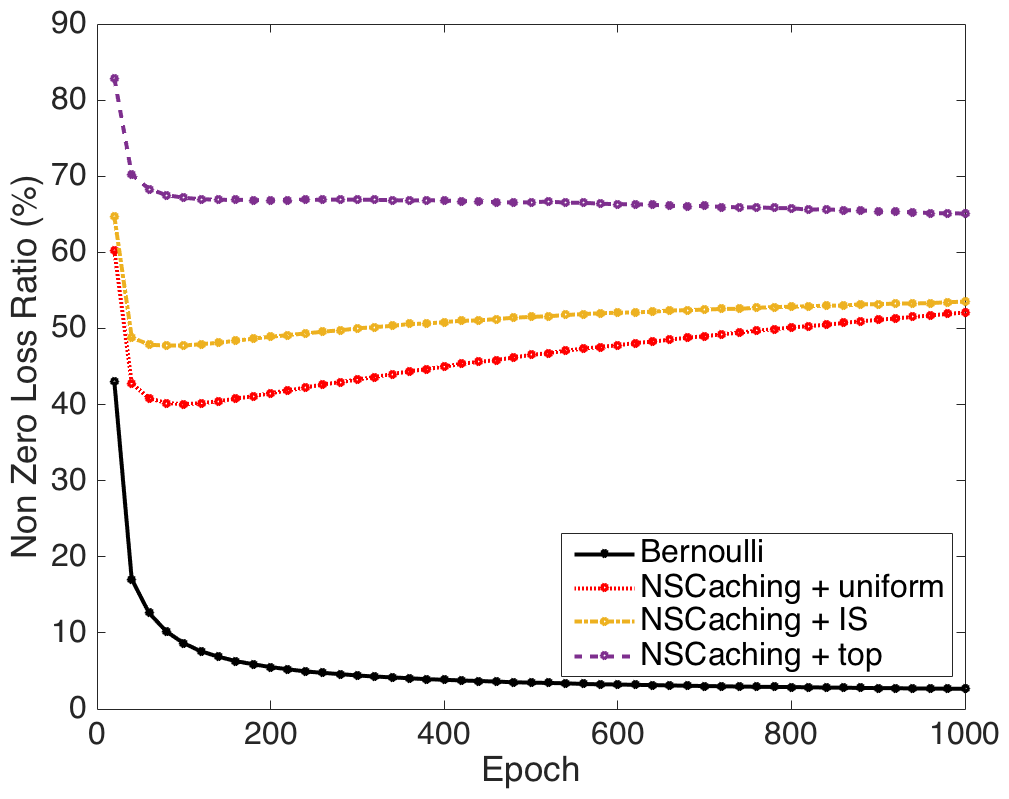}
\label{fig:sampleee:loss}}

\caption{Balancing on exploration (left) and exploitation (right) of different sampling strategies. Evaluated by TransD model on WN18.}
\label{fig:sampleee}
\end{figure}

The RR is shown in Figure \ref{fig:sampleee:repeat}. 
The Bernoulli sampling method has almost zero repeat triplets since the number of explored negatives is extremely large, 
it has the best exploration. 
Among the schemes based on NSCaching, uniform sampling has better exploration than IS, 
then followed by top sampling. NZL ratio is shown in Figure \ref{fig:sampleee:loss}. 
{
As training going on, the baseline Bernoulli model suffers the zero loss problem severely, thus leading to vanishing gradient.
}
All of the three schemes have more than half non-zero losses, thus achieves exploitation.
To sum up,
uniform sampling is the most balanced strategy among 
the three schemes,
thus \textit{NSCaching + uniform} achieving the best performance.

\subsubsection{Importance sampling strategy to update the cache (step~8)}

As discussed in Section~\ref{sssec:update},
we have two choices on updating the cache:
(i) importance sampling based method, which samples $N_1$ entities from $N_1+N_2$ candidates according to the probability in (\ref{eq:prupc}) without replacement, (IS update).
(ii) top sampling method, which directly select $N_1$ entities with top scores in the candidates, (top update).
Again,
let us first look at performance comparison in 
Figure~\ref{fig:ablation:sample}.(b).
We can see that IS update outperforms top update by a large margin.

%

Then,
to explain the exploration and exploitation here,
we add two extra measurements for comparison.
They are (i). the number of changed elements in cache (denoted as ``CE'')
and (ii) the ratio of non-zero losses, i.e., NZL.
More changed elements leads to larger exploration, 
and more nonzero losses means more exploitation.

\begin{figure}[ht]
\subfigure[CE v.s. epoch.]
{\includegraphics[width=0.235\textwidth]{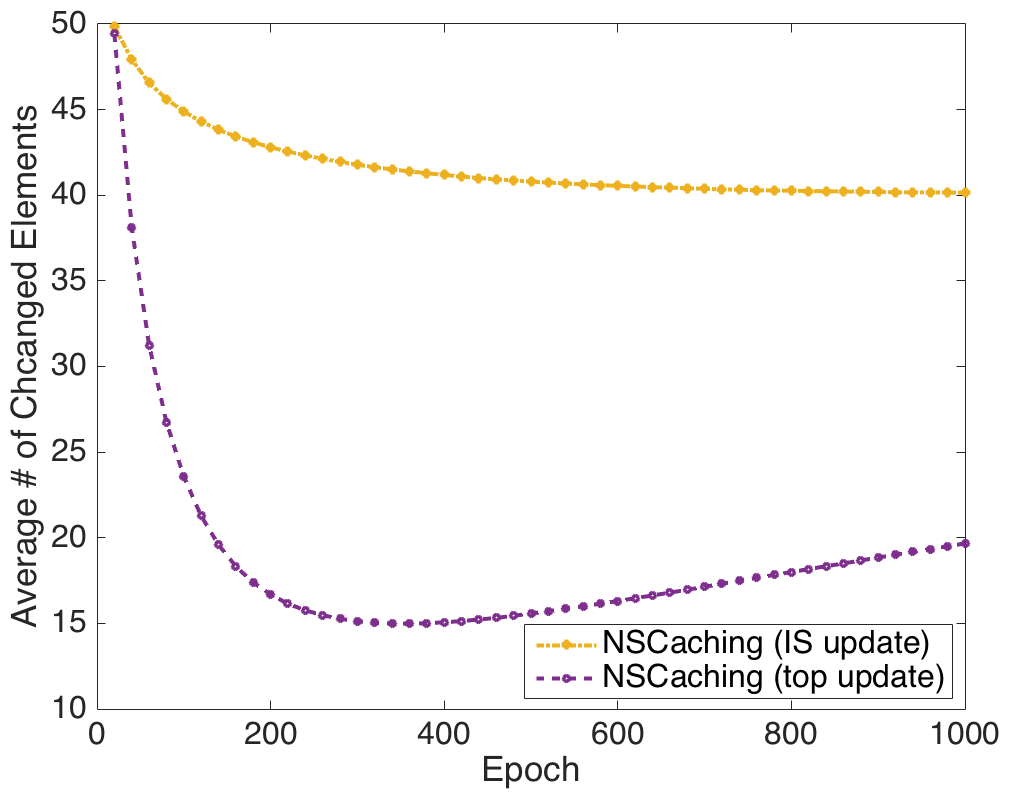} \label{fig:update:change}}
\subfigure[NZL v.s. epoch.]
{\includegraphics[width=0.234\textwidth]{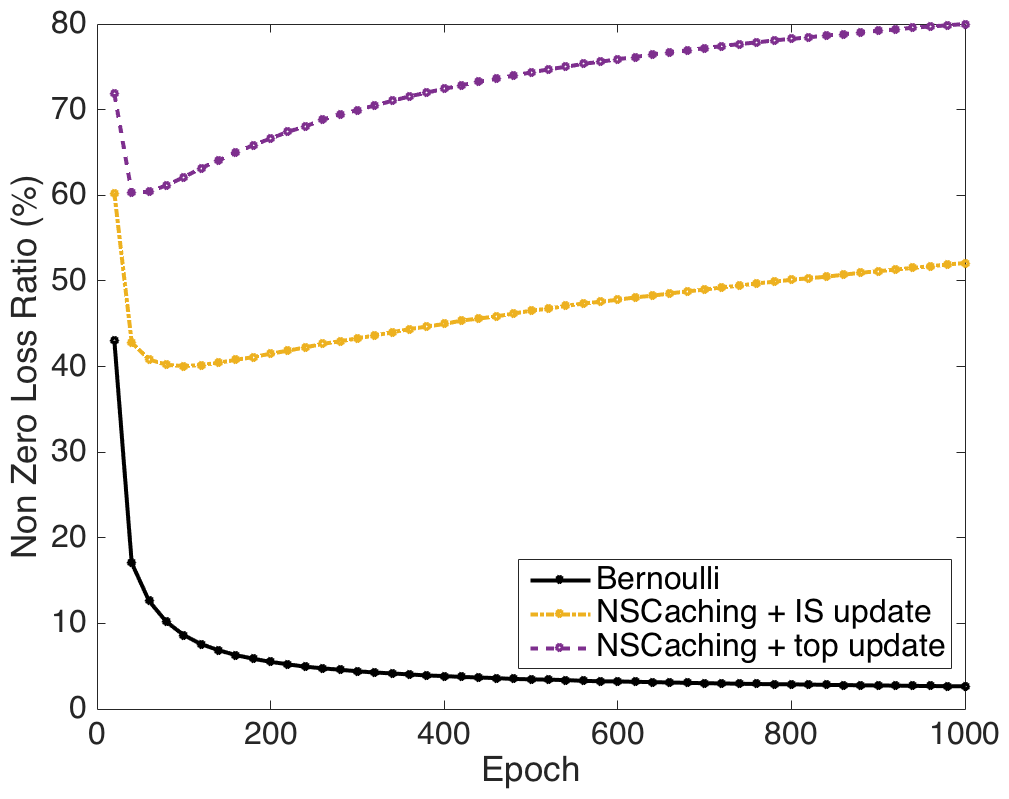} \label{fig:update:loss}}

\caption{Balancing on exploration (left) and exploitation (right) of different strategies for updating the cache. Evaluated by TransD model on WN18.}
\label{fig:update:ee}
 \end{figure}

The value of CE measures the different elements in the cache in a period of epochs. 
As shown in Figure \ref{fig:update:ee}.(a), 
the number of changed elements in top update scheme is much smaller than that of the importance sampling update. 
As a result, the cache is updated quite slow and the model mainly focuses on these highly scored negative triplets,
 which may contain many false positive triplets.
 As a comparison, the importance sampling based update scheme can keep the cache fresh and keep track of dynamic changes of the negative sampling distribution. 
 It not only provides enough qualified negative triplets for the KG embedding model to avoid zero loss, 
 but also explore the large negative sample space well. In summary, we choose the importance sampling strategy to update the cache.

\subsection{Sensitivity Analysis: Cache Size}
\label{ssec:exp:size}

Comparing with the baseline KG embedding models (i.e., Bernoulli \cite{wang2014knowledge,lin2015learning}),
the only extra hyper-parameters here are $N_1$ and $N_2$.
Basically,
$N_1$ is the size of cache.
Then,
$N_2$ is the size of randomly sampled negative triplets from $\bar{\mathcal{S}}_{(h,r,t)}$,
which will be later used to update the cache.
Here,
we show their impact on NSCaching's performance.

\begin{figure}[ht]
\centering
\subfigure[Diff. $N_1$]
{\includegraphics[width = 0.238\textwidth]{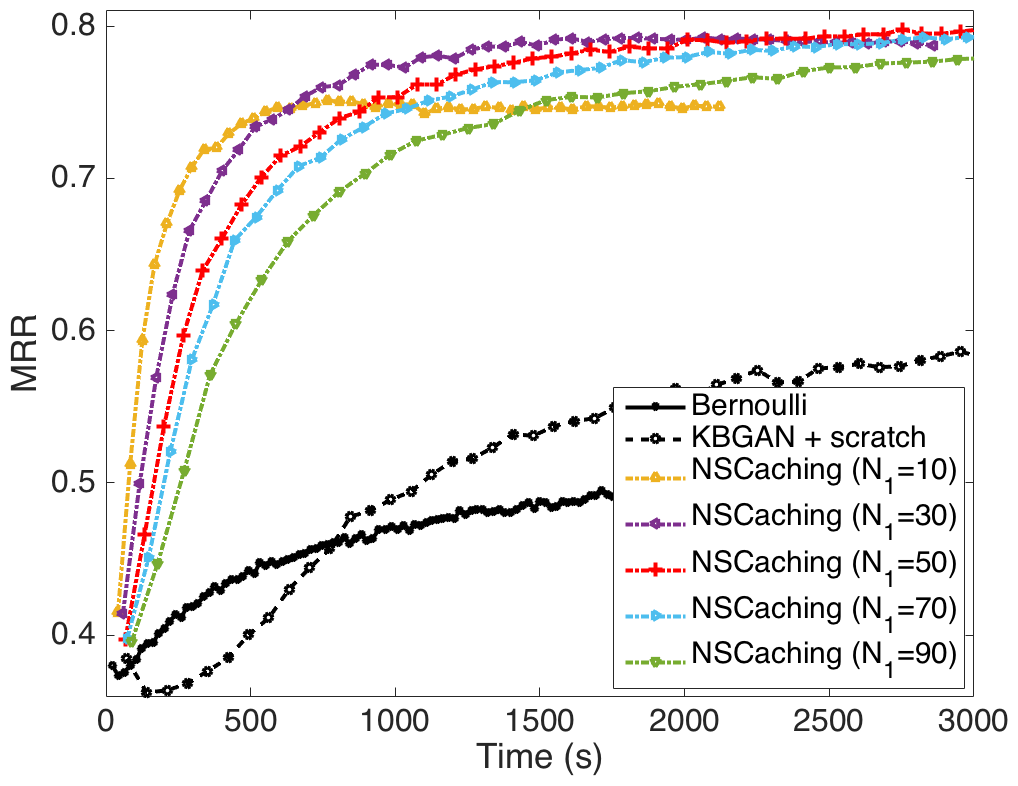}}
\subfigure[Diff. $N_2$]
{\includegraphics[width = 0.235\textwidth]{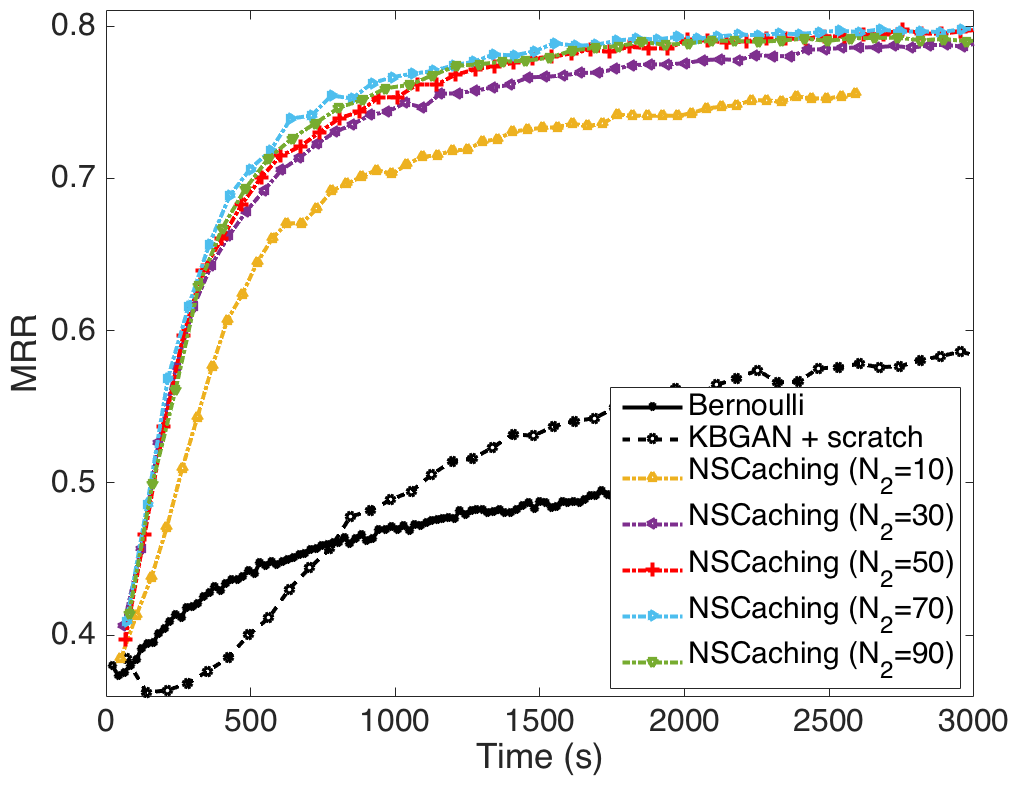}}

\caption{Comparison of different $N_1$ when random subset size $N_2$ is fixed to 50, and different $N_2$ when cache size $N_1$ is fixed to 50. Evaluated by TransD model on WN18.}
\label{fig-CS-wn18-transd}
\end{figure}

Figure~\ref{fig-CS-wn18-transd}.(a) shows how performance changes by varying the cache size $N_1$ among $\{10, 30, 50, 70, 90\}$, with fixed $N_2=50$. 
When the cache size is small, average score of entities stored in cache should be larger than those in larger cache. 
Thus, 
false negative samples will be more likely to be sampled, which will influence the boundary to a bad location.
As for the others values of $N_1$, NSCaching performs quite stable. 
The convergence speed is similar, as well as the values in converged state.
Thus, when finding appropriate cache size, the value of $N_1$ can be searched from smaller value until the performance is stable.

%

%

Different performance of the random candidate subset size $N_2$ is shown in Figure~\ref{fig-CS-wn18-transd}.(b).
Obviously, the entities in cache will be updated more frequently when $N_2$ gets larger, 
which lead to better exploration. But the trade-off is that larger value of $N_2$ costs more. 
As shown by the colored lines in Figure \ref{fig-CS-wn18-transd}.(b), NSCaching performs consistently when $N_2$ is larger than 10.
However, if the random subset is small, the content in cache will be harder to be updated, thus lead to poor performance as the yellow dashed line ($N_2=10$).

By combining together the influence of cache size $N_1$ in Figure \ref{fig-CS-wn18-transd} and random subset size $N_2$ in Figure \ref{fig-CS-wn18-transd}.(b),
we find that 
(i) NSCaching is not sensitive to the two sizes; 
(ii) both sizes can not be too small; 
(iii) $N_1=N_2$ is a good balance.

{
\subsection{Illustration of Vanishing Gradient}

To further clarity the vanishing gradient problem, we plot average $\ell_2$-norm of gradients v.s. number of epochs in Figure~\ref{fig-GN-wn18rr}. 
Note that Adam \cite{kingma2014adam}, which is a stochastic gradient descent algorithm, is used as the optimizer.
First, we can see that while norms of gradients for both \textit{NSCaching} and \textit{Bernoulli} become smaller, 
they will not decrease to zero since the sampling process of the mini-batch will introduce noisy into gradients.
However, the norm from \textit{NSCaching} is larger than that from \textit{Bernoulli},
which dues to the usage of caching-based negative sampling scheme.
Thus, we can see \textit{NSCaching} can successfully avoid the problem of vanishing gradient.

\begin{figure}[ht]
\centering
\subfigure[TransD]
{\includegraphics[width = 0.238\textwidth]{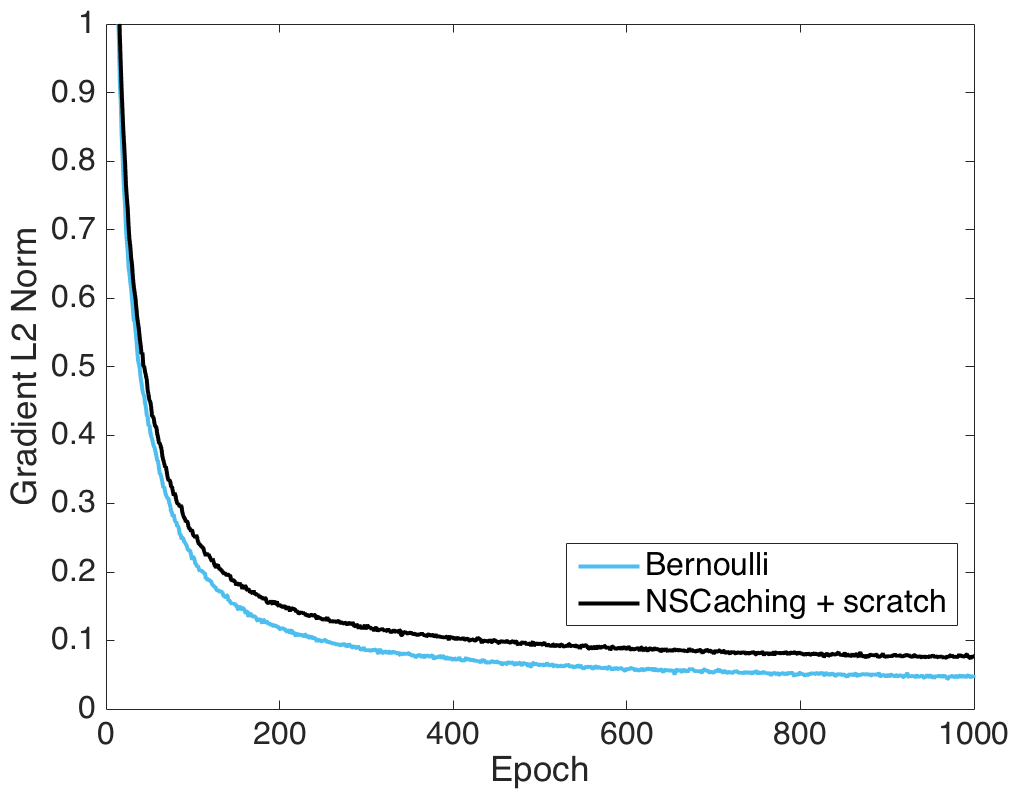}}
\subfigure[ComplEx]
{\includegraphics[width = 0.238\textwidth]{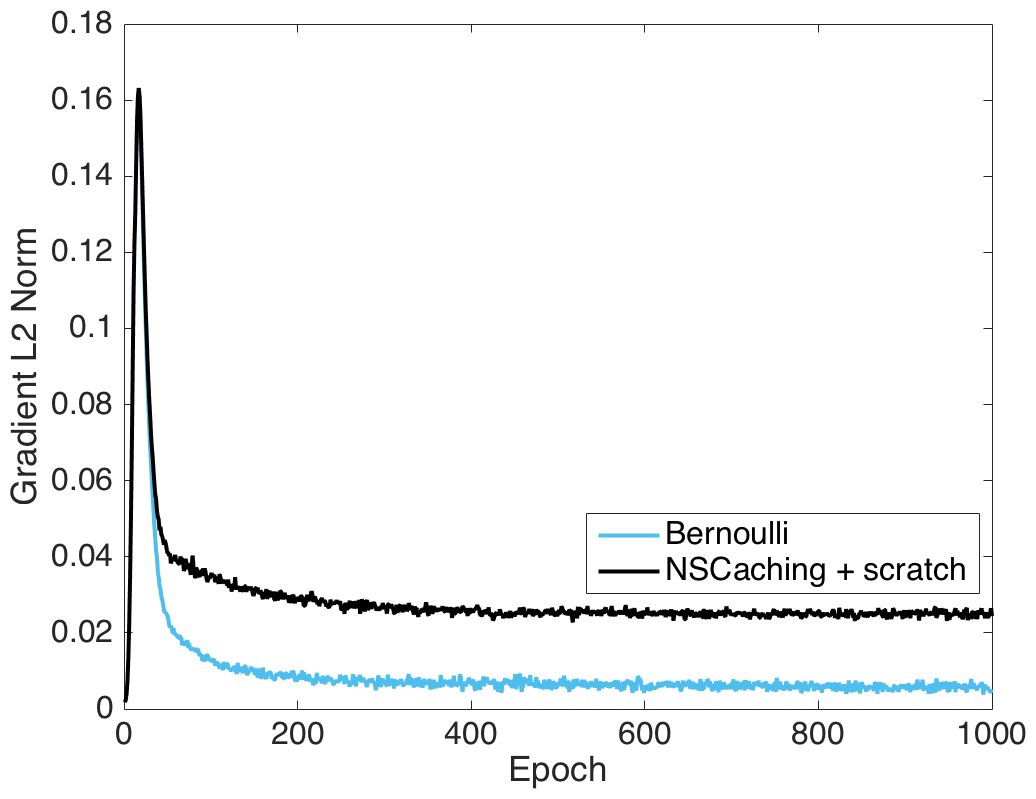}}

\caption{Mini-batch's average $\ell_2$-norm of gradients in one epoch v.s. number of epochs for Bernoulli and NSCaching on WN18RR.}
\label{fig-GN-wn18rr}
\end{figure}
}

\subsection{Explanation of the connection to Self-Paced Learning}
\label{sec:visself}

Finally,
we visualize the changes of entities in the cache,
which verifies the effects of self-paced learning introduced in Section~\ref{sec:self-pace}. 
Following \cite{wang2018incorporating},
we also use FB13 here since its triplets are more interpretable than the four evaluated datasets.
We pick up 
$(manorama$, $profession$, $actor)$ as the positive triplets, 
and the changes in its tail-cache are show in Table~\ref{tab:cache:example}.
As can be seen,
entities are firstly meaninglessness,
e.g., \textit{ostrava} and \textit{ben\_lilly},
then they gradually changes to human jobs,
e.g., \textit{artist} and \textit{sex\_worker}.

\begin{table}[ht]
	\centering
	\caption{Examples of negative entities in cache on FB13. Each line displays 5 random sampled negative entities from tail cache of a positive fact \textit{(manorama, profession, actor)} in different epochs.}
	\label{tab:cache:example}
	\begin{tabular}{c|c}
		\hline 
		epoch & entities in cache \\ \hline
		0 & \textit{allen\_clarke,  jose\_gola, ostrava, ben\_lilly, hans\_zinsser} \\ \hline
		20 & \textit{accountant, frank\_pais, laura\_marx, como, domitia\_lepida}   \\ \hline
		100 & \textit{artist, , aviator, hans\_zinsse, john\_h\_cough}    \\ \hline
		200 & \textit{physician, artist, raich\_carter, coach, mark\_shivas}      \\ \hline
		500 & \textit{artist, physician, cavan, sex\_worker, attorney\_at\_law} \\ \hline
	\end{tabular}
\end{table}


\section{Related work}
\label{sec:rel-work}
\subsubsection{Generative Adversarial Network} 
Generative Adversarial Network (GAN) is originally introduced as a powerful model for plausible image generation. 
The GAN contains two modules: 
a \emph{generator} that serves as a complex distribution sampler, 
and a \emph{discriminator} that measures the quality of generated samples. 
Under elaborately control on the training procedure of generator and discriminator \cite{arjovsky2017wasserstein,gulrajani2017improved}, 
GAN achieved significant success computer vision field \cite{zhang2018self,zhu2017unpaired}. 
It has been shown to sample high-quality negative samples for knowledge graph embedding \cite{cai2018kbgan,wang2018incorporating}.

\subsubsection{Negative Sampling} 
Negative sampling is originally introduced as an alternative to the hierarchical softmax, 
which aims at reducing complexity of softmax on large scale dataset \cite{gutmann2010noise}. 
It then becomes popular in embedding learning, especially for 
word embedding \cite{goldberg2014word2vec}, graph embedding \cite{grover2016node2vec}, 
and KG embedding \cite{wang2017knowledge}.
More 
recently, there have been interests in applying the GAN to negative sampling,
e.g., IGAN \cite{wang2018incorporating} and KBGAN \cite{cai2018kbgan} for KG embedding
and self-paced GAN \cite{gao2018self} for network embedding.

\section{Conclusion}
\label{sec:conclude}

We proposed NSCaching as a novel negative sampling method for knowledge graph embedding learning. 
The negative samples are from a cache that can dynamically hold high-quality negative samples. 
We analyze the designing of NSCaching through the balance of exploration and exploitation. 
Experimentally, we empirically test NSCaching on two datasets and five scoring functions. 
Results show that the method can generalize well under various settings and achieves state-of-the-arts performance on FB15K dataset.
When dealing with millions scale KG, memory of storing the cache becomes a problem. 
Using distributed computation or hashing will be pursued as future works. 
{Besides,
	the theoretical convergence of NSCaching is also an important and interesting future work.}



\bibliographystyle{plain}
\bibliography{bib}

\begin{thebibliography}{10}

\bibitem{arjovsky2017wasserstein}
M.~Arjovsky, S.~Chintala, and L.~Bottou.
\newblock Wasserstein {GAN}.
\newblock Technical report, 2017.

\bibitem{auer2007dbpedia}
S.~Auer, C.~Bizer, G.~Kobilarov, J.~Lehmann, R.~Cyganiak, and Z.~Ives.
\newblock {DB}pedia: A nucleus for a web of open data.
\newblock In {\em The Semantic Web}, pages 722--735. Springer, 2007.

\bibitem{bengio2009curriculum}
Y.~Bengio, J.~Louradour, R.~Collobert, and J.~Weston.
\newblock Curriculum learning.
\newblock In {\em ICML}, pages 41--48. ACM, 2009.

\bibitem{bollacker2008freebase}
K.~Bollacker, C.~Evans, P.~Paritosh, T.~Sturge, and J.~Taylor.
\newblock Freebase: a collaboratively created graph database for structuring
  human knowledge.
\newblock In {\em SIGMOD}, pages 1247--1250. ACM, 2008.

\bibitem{bordes2014question}
A.~Bordes, S.~Chopra, and J.~Weston.
\newblock Question answering with subgraph embeddings.
\newblock In {\em EMNLP}, pages 615--620, 2014.

\bibitem{bordes2014semantic}
A.~Bordes, X.~Glorot, J.~Weston, and Y.~Bengio.
\newblock A semantic matching energy function for learning with
  multi-relational data.
\newblock {\em Machine Learning}, 94(2):233--259, 2014.

\bibitem{bordes2013translating}
A.~Bordes, N.~Usunier, A.~Garcia-Duran, J.~Weston, and O.~Yakhnenko.
\newblock Translating embeddings for modeling multi-relational data.
\newblock In {\em NIPS}, pages 2787--2795, 2013.

\bibitem{bordes2014open}
A.~Bordes, J.~Weston, and N.~Usunier.
\newblock Open question answering with weakly supervised embedding models.
\newblock In {\em ECML-PKDD}, pages 165--180. Springer, 2014.

\bibitem{cai2018kbgan}
L.~Cai and W.Y. Wang.
\newblock Kbgan: Adversarial learning for knowledge graph embeddings.
\newblock In {\em ACL}, volume~1, pages 1470--1480, 2018.

\bibitem{drumond2012predicting}
L.~Drumond, S.~Rendle, and L.~Schmidt-Thieme.
\newblock Predicting rdf triples in incomplete knowledge bases with tensor
  factorization.
\newblock In {\em SAC}, pages 326--331, 2012.

\bibitem{fan2014transition}
M.~Fan, Q.~Zhou, E.~Chang, and T.~F. Zheng.
\newblock Transition-based knowledge graph embedding with relational mapping
  properties.
\newblock In {\em PACLIC}, 2014.

\bibitem{gao2018self}
H.~Gao and H.~Huang.
\newblock Self-paced network embedding.
\newblock In {\em SIGKDD}, pages 1406--1415, 2018.

\bibitem{getoor2007introduction}
L.~Getoor and B.~Taskar.
\newblock {\em Introduction to statistical relational learning}, volume~1.
\newblock The MIT Press, 2007.

\bibitem{glorot2010understanding}
X.~Glorot and Y.~Bengio.
\newblock Understanding the difficulty of training deep feedforward neural
  networks.
\newblock In {\em AISTATS}, pages 249--256, 2010.

\bibitem{goldberg2014word2vec}
Y.~Goldberg and O.~Levy.
\newblock word2vec explained: deriving mikolov et al.'s negative-sampling
  word-embedding method.
\newblock Technical report, 2014.

\bibitem{goodfellow2014generative}
I.~Goodfellow, J.~Pouget-Abadie, M.~Mirza, B.~Xu, D.~Warde-Farley, S.~Ozair,
  A.~Courville, and Y.~Bengio.
\newblock Generative adversarial nets.
\newblock In {\em NIPS}, pages 2672--2680, 2014.

\bibitem{grover2016node2vec}
A.~Grover and J.~Leskovec.
\newblock node2vec: Scalable feature learning for networks.
\newblock In {\em SIGKDD}, pages 855--864. ACM, 2016.

\bibitem{gulrajani2017improved}
I.~Gulrajani, F.~Ahmed, M.~Arjovsky, V.~Dumoulin, and A.~C. Courville.
\newblock Improved training of wasserstein gans.
\newblock In {\em NIPS}, pages 5767--5777, 2017.

\bibitem{gutmann2010noise}
M.~Gutmann and A.~Hyv{\"a}rinen.
\newblock Noise-contrastive estimation: A new estimation principle for
  unnormalized statistical models.
\newblock In {\em AISTATS}, pages 297--304, 2010.

\bibitem{ji2015knowledge}
G.~Ji, S.~He, L.~Xu, K.~Liu, and J.~Zhao.
\newblock Knowledge graph embedding via dynamic mapping matrix.
\newblock In {\em ACL}, volume~1, pages 687--696, 2015.

\bibitem{ji2016knowledge}
G.~Ji, K.~Liu, S.~He, and J.~Zhao.
\newblock Knowledge graph completion with adaptive sparse transfer matrix.
\newblock In {\em AAAI}, pages 985--991, 2016.

\bibitem{kingma2014adam}
Diederik~P Kingma and Jimmy Ba.
\newblock Adam: A method for stochastic optimization.
\newblock Technical report, 2014.

\bibitem{kok2007statistical}
S.~Kok and P.~Domingos.
\newblock Statistical predicate invention.
\newblock In {\em ICML}, pages 433--440, 2007.

\bibitem{kumar2010self}
M.~P. Kumar, B.~Packer, and D.~Koller.
\newblock Self-paced learning for latent variable models.
\newblock In {\em NIPS}, pages 1189--1197, 2010.

\bibitem{lao2011random}
N.~Lao, T.~Mitchell, and W.~W. Cohen.
\newblock Random walk inference and learning in a large scale knowledge base.
\newblock In {\em EMNLP}, pages 529--539. Association for Computational
  Linguistics, 2011.

\bibitem{lin2015learning}
Y.~Lin, Z.~Liu, M.~Sun, Y.~Liu, and X.~Zhu.
\newblock Learning entity and relation embeddings for knowledge graph
  completion.
\newblock In {\em AAAI}, volume~15, pages 2181--2187, 2015.

\bibitem{liu2017analogical}
H.~Liu, Y.~Wu, and Y.~Yang.
\newblock Analogical inference for multi-relational embeddings.
\newblock In {\em ICML}, pages 2168--2178, 2017.

\bibitem{mikolov2013linguistic}
T.~Mikolov, W.~Yih, and G.~Zweig.
\newblock Linguistic regularities in continuous space word representations.
\newblock In {\em ACL}, pages 746--751, 2013.

\bibitem{miller1995wordnet}
G.~A. Miller.
\newblock Wordnet: a lexical database for english.
\newblock {\em Communications of the {ACM}}, 38(11):39--41, 1995.

\bibitem{nickel2016review}
M.~Nickel, K.~Murphy, V.~Tresp, and E.~Gabrilovich.
\newblock A review of relational machine learning for knowledge graphs.
\newblock {\em Proceedings of the IEEE}, 104(1):11--33, 2016.

\bibitem{nickel2016holographic}
M.~Nickel, L.~Rosasco, and T.~A. Poggio.
\newblock Holographic embeddings of knowledge graphs.
\newblock In {\em AAAI}, volume~2, pages 3--2, 2016.

\bibitem{nickel2011three}
M.~Nickel, V.~Tresp, and H.~Kriegel.
\newblock A three-way model for collective learning on multi-relational data.
\newblock In {\em ICML}, volume~11, pages 809--816, 2011.

\bibitem{paszke2017automatic}
A.~Paszke, S.~Gross, S.~Chintala, G.~Chanan, E.~Yang, Z.~DeVito, Z.~Lin,
  A.~Desmaison, L.~Antiga, and A.~Lerer.
\newblock Automatic differentiation in pytorch.
\newblock Technical report, 2017.

\bibitem{singhal2012introducing}
A.~Singhal.
\newblock Introducing the knowledge graph: things, not strings.
\newblock {\em Official Google blog}, 5, 2012.

\bibitem{zhang2018self}
Q.~Song, H.~Ge, J.~Caverlee, and X.~Hu.
\newblock Self-attention generative adversarial networks.
\newblock Technical report, 2018.

\bibitem{suchanek2007yago}
F.~M. Suchanek, G.~Kasneci, and G.~Weikum.
\newblock Yago: a core of semantic knowledge.
\newblock In {\em WWW}, pages 697--706, 2007.

\bibitem{toutanova2015observed}
Kristina Toutanova and Danqi Chen.
\newblock Observed versus latent features for knowledge base and text
  inference.
\newblock In {\em Workshop on Continuous Vector Space Models and their
  Compositionality}, pages 57--66, 2015.

\bibitem{trouillon2016complex}
T.~Trouillon, J.~Welbl, S.~Riedel, and G.~Gaussier, {\'E}.
\newblock Complex embeddings for simple link prediction.
\newblock In {\em ICML}, pages 2071--2080, 2016.

\bibitem{wang2018incorporating}
P.~Wang, S.~Li, and R.~Pan.
\newblock Incorporating {GAN} for negative sampling in knowledge representation
  learning.
\newblock {\em AAAI}, 2018.

\bibitem{wang2017knowledge}
Q.~Wang, Z.~Mao, B.~Wang, and L.~Guo.
\newblock Knowledge graph embedding: A survey of approaches and applications.
\newblock {\em TKDE}, 29(12):2724--2743, 2017.

\bibitem{wang2018evaluating}
Y.~Wang, D.~Ruffinelli, S.~Broscheit, and R.ainer Gemulla.
\newblock On evaluating embedding models for knowledge base completion.
\newblock {\em arXiv preprint arXiv:1810.07180}, 2018.

\bibitem{wang2014knowledge}
Z.~Wang, J.~Zhang, J.~Feng, and Z.~Chen.
\newblock Knowledge graph embedding by translating on hyperplanes.
\newblock In {\em AAAI}, volume~14, pages 1112--1119, 2014.

\bibitem{white1991knowledge}
D.~A. White.
\newblock The knowledge-based software assistant: A program summary.
\newblock In {\em ICKBSE}, pages 2--6, 1991.

\bibitem{williams1992simple}
R.~J. Williams.
\newblock Simple statistical gradient-following algorithms for connectionist
  reinforcement learning.
\newblock {\em Machine Learning}, 8(3-4):229--256, 1992.

\bibitem{xiao2016from}
H.~Xiao, M.~Huang, and X.~Zhu.
\newblock From one point to a manifold: knowledge graph embedding for precise
  link prediction.
\newblock In {\em IJCAI}, pages 1315--1321, 2016.

\bibitem{yang2014embedding}
B.~Yang, W.~Yih, X.~He, J.~Gao, and L.~Deng.
\newblock Embedding entities and relations for learning and inference in
  knowledge bases.
\newblock Technical report, 2017.

\bibitem{zhu2017unpaired}
J.~Zhu, T.~Park, P.~Isola, and A.~A. Efros.
\newblock Unpaired image-to-image translation using cycle-consistent
  adversarial networks.
\newblock In {\em ICCV}, pages 2242--2251. IEEE, 2017.

\end{thebibliography}

\end{document}